\pdfoutput=1

\documentclass[final,3p,times]{elsarticle}

\usepackage{graphicx}
\usepackage[fleqn]{amsmath}
\usepackage{amsfonts}
\usepackage{amssymb}
\usepackage{amsthm}
\usepackage{thmtools}
\allowdisplaybreaks

\usepackage[algoruled,titlenotnumbered]{algorithm2e}
\usepackage{wrapfig}
\usepackage{natbib}
\usepackage{multirow}
\usepackage{graphicx}
\usepackage{hyperref}
\hypersetup{
    colorlinks=true,
    citecolor=blue,
    linkcolor=red,
    filecolor=magenta,      
    urlcolor=magenta,
    }

\usepackage{booktabs}
\usepackage{multirow}
\usepackage{float}
\usepackage{subfig}
\usepackage{wrapfig}
\usepackage{algpseudocode}
\usepackage[export]{adjustbox}
\usepackage{bbm}
\usepackage{afterpage}
\usepackage[dvipsnames,table]{xcolor}

\newcommand{\RR}{\mathbb{R}}

\newcommand{\PP}{\mathcal{P}}
\newcommand{\QQ}{\mathcal{Q}}
\newcommand{\FF}{\mathcal{F}}

\newcommand{\GX}{G}

\newcommand{\LL}{\mathcal{L}}
\newcommand{\RX}{\mathcal{R}}

\renewcommand{\aa}{\mathbf{a}}
\newcommand{\uu}{\mathbf{u}}
\newcommand{\zz}{\mathbf{z}}
\newcommand{\qq}{\mathbf{q}}

\newcommand{\fno}{\text{FNO}}
\newcommand{\pifno}{\pi\text{-}\text{FNO}}

\newcommand{\ZZ}{\mathbb{Z}}

\journal{Elsevier}
\biboptions{authoryear}
\graphicspath{{./Figs/}}

\begin{document}
	
\begin{frontmatter}
	
    \title{Fourier neural operator for learning solutions to macroscopic traffic flow models: Application to the forward and inverse problems}
    
    \author[1,2]{Bilal Thonnam Thodi}
    \author[2]{Sai Venkata Ramana Ambadipudi}
    \author[1,2]{Saif Eddin Jabari\corref{cor1}}
    \cortext[cor1]{Corresponding author, e-mail: \url{sej7@nyu.edu}}
    \address[1]{New York University Tandon School of Engineering, Brooklyn, NY 12011, U.S.A.}
    \address[2]{New York University Abu Dhabi, Saadiyat Island, P.O. Box 129188, Abu Dhabi, U.A.E.}
		


    \begin{abstract}
    Deep learning methods are emerging as popular computational tools for solving forward and inverse problems in traffic flow. In this paper, we study a neural operator framework for learning solutions to first-order macroscopic traffic flow models with applications in estimating traffic densities for urban arterials. In this framework, an operator is trained to map heterogeneous and sparse traffic input data to the complete macroscopic traffic density in a supervised learning setting. We chose a physics-informed Fourier neural operator ($\pi$-FNO) as the operator, where an additional physics loss based on a discrete conservation law regularizes the problem during training to improve the shock predictions. We also propose to use training data generated from random piecewise constant input data to systematically capture the shock and rarefaction solutions of certain macroscopic traffic flow models. From experiments using the LWR traffic flow model, we found superior accuracy in predicting the density dynamics of a ring-road network and urban signalized road. We also found that the operator can be trained using simple traffic density dynamics, e.g., consisting of $2-3$ vehicle queues and $1-2$ traffic signal cycles, and it can predict density dynamics for heterogeneous vehicle queue distributions and multiple traffic signal cycles $(\geq 2)$ with an acceptable error. The extrapolation error grew sub-linearly with input complexity for a proper choice of the model architecture and training data. 
    Adding a physics regularizer aided in learning long-term traffic density dynamics, especially for problems with periodic boundary data.
    \end{abstract}
		
		
\end{frontmatter}
	
	
\section{Introduction}
\label{sec:intro}

Physics-informed machine learning (PIML) methods that integrate data-driven algorithms with physical priors or domain knowledge have recently drawn significant interest in scientific computing applications,
and traffic flow is no exception. Some recent applications of PIML in traffic include flow modeling and simulations \citep{thodi2023_thesis,thodi2022pifno,zhang2022cfm_gan,mo2021pidl_cfm,yuan2021macro_gp}, state estimation \citep{xuan2023physgan_traffic,chuhan2023ifac,thodi2022aniso,thodi2021aniso_itsc,shi2021pinns_tse,huang2020pidl_tse,jabari2019learning}, traffic predictions \citep{li2022pinns_networkpred,pereira2022rnn_phys, liu2021tse}, and traffic control \citep{han2022pinn_ramp,di2021survey_aicontrol}.  Of particular interest are the deep learning methods for solving forward and inverse problems that involve partial differential equations (PDE) that arise as a key component in most real-time traffic simulation and estimation methods \citep{kessel2019tfm}. 

There are two broad categories of deep learning-based PDE solvers studied in the literature: physics-informed neural networks (PINNs) \citep{raissi2019pinns} and neural operators (NOs) \citep{ lu2021deeponet, li2021fno}. 
PINNs use deep neural networks to approximate the PDE solution, which is trained to minimize a physics loss $-$ PDE residual evaluated at random collocation points in the domain and a data loss $-$ solution error at the given input or boundary data. On the other hand, NOs have shown that the underlying solution mapping can be learned even without knowledge of the PDE but with historical simulation data in a supervised learning framework.  Both methods offer advantages over traditional PDE solvers in terms of handling sparse and heterogeneous input data, grid-independent solutions, and computational cost. Despite the improvements, deep learning-based solvers can still generate unrealistic or non-physical predictions owing to a lack of physical information about the problem during training and a poor choice of data for training the solvers \citep{wang2021gradientflow}.

\begin{figure}[tb!]
    \centering
    \includegraphics[width=0.95\textwidth]{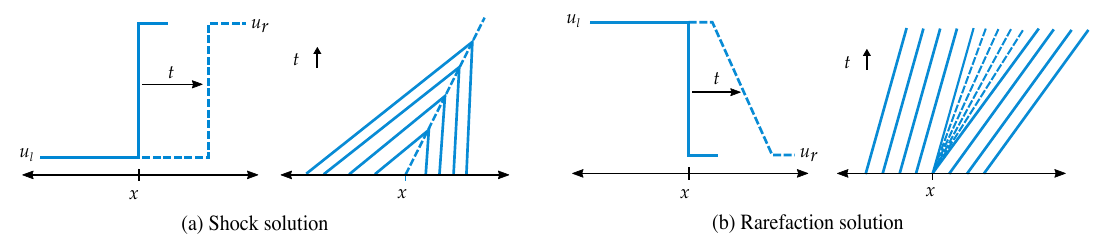}
    \caption{Two types of Riemann density solutions possible for LWR-type traffic flow models. The solutions correspond to the condition (a)  $u_{l} \leq u_{r} < u_{\rm cr}$ and (a)  $u_{r} \leq u_{l} < u_{cr}$, where $u_{\rm cr} = \max_{u} f(u)$ is the critical traffic density.}
    \label{fig:riemann}
\end{figure}

We consider the Lighthill-Whitham and Richards (LWR PDE) traffic flow model \citep{light1955lwr,richards1956lwr}, $\partial_{t} u + f'(u) \partial_{x} u = 0$, where $u$ denotes traffic density with concave flux function $f$ (the fundamental diagram). The wave motion in the LWR PDE is characterized by finite propagation speeds $f'(u)$. The solutions are characteristic trajectories emanating from the datum (e.g., the initial conditions).  As the wave speed $f'(u)$ is density-dependent,  the characteristic lines may cross somewhere in the domain within a finite time. At these domain points, the solution is multi-valued, which means that the problem does not have a well-posed solution in the classical sense. Hence, weak solutions formed by a combination of shocks (discontinuities or kinks) and rarefactions (continuously varying densities) are introduced \citep{whitham1999waves, leveque1992numerical}. An example of shock and rarefaction solutions is illustrated in Figure \ref{fig:riemann}.

We are interested in learning solutions to forward and inverse problems that involve the LWR PDE for applications in traffic flow. Numerical solvers for the forward problem are well-studied in the transportation literature: various classical solvers have been tailored to solving the LWR PDE, including finite difference techniques \citep{michalopoulos1984finite}, finite elements \citep{beskos1984finite}, and finite volume methods \citep{daganzo1994ctm,daganzo1995ctm_net,lebacque1996godunov}. Other classical techniques have been tailored to solving Hamilton-Jacobi and Lagrangian variants of the LWR PDE \citep{daganzo2005var,claudel2010laxhopf,mazare2011analytical,leclercq2007lagrang}. In the class of inverse problems, we consider traffic state estimation problems, where sparse measurements from point sensors or mobile sensors are available, and the objective is to reconstruct traffic variables throughout the spatio-temporal domain of interest. Examples of classical techniques that were developed for these kinds of inverse problems in the transportation literature include \citep{blandind2012tse_mfd,hoogen2012lagrang,jabari2013gauss,canepa2017network_hjb,zheng2018stochastic}.  Recent traffic-related works explored the use of physics-informed neural networks (PINNs) for both the forward and the inverse problems \citep{xuan2023physgan_traffic,shi2021pinns_tse,liu2021tse,rempe20221tse_pidl,huang2022pinns_lwr,huang2020pidl_tse}. 
However, some limitations are associated with PINNs.

PINNs have shown successful results in solving several forward and inverse problems with underlying solutions that are smooth. However, PINNs have only seen limited success in approximating solutions having high irregularities, such as shock solutions of non-linear hyperbolic conservation laws, which include the LWR PDE  \citep{patel2022cvpinns,de2022weak_pinns,jagtap2020cpinns}. Due to possible non-differentiabilities (``kinks'') in the solution, the partial derivatives $\partial_{t} u$ and $\partial_{x} u$ are not well-defined at all points in the computational domain. This implies that the physics loss function used for training neural networks is not well-defined and leads to poor convergence results. \cite{huang2023pinns_limitation} raised this concern while solving the LWR traffic flow model.  Even for viscous regularized versions, where a viscosity term is added to smoothen discontinuities and kinks, the derivatives can have high variations near the jumps, and residuals can blow up in those regions. Also, neural networks are theoretically shown to poorly approximate discontinuous or highly varying functions. Typically, the approximation errors for deep neural networks scale in the order $\sim O \big( \epsilon^{-d/n} \big)$, for a desirable approximation error $\epsilon$ \citep{yarotsky2017error_deeprelu}. $d$ and $n$ are the dimension and smoothness of the approximating function class; larger $n$ implies a smooth function and vice-versa. This means highly varying functions (i.e., $n \rightarrow 0$) require exponentially large neural networks to ensure a lower error $\epsilon$.

A few recent attempts focused on overcoming the above challenges of learning highly-varying solutions. For instance, \citep{jagtap2020cpinns,dwivedi2021pielm_distributed} proposes to divide the computational domain into multiple sub-domains, each approximated with different neural networks. This allows for parallel computing, where the approximation task is distributed across multiple, presumably smaller-size, neural networks. \citep{wu2023pinns_sampling,mao2020pinns_highspeed} proposes an adaptive sampling method to choose larger collocation points near the regions of high irregularities, thus giving more weight to the discontinuous region in the loss function. In the context of traffic, \citep{huang2023pinns_limitation, shi2021pinns_tse} propose to solve the viscous regularized LWR PDE to ensure a well-posed physics loss function.
Despite their practical success, these guided improvements have several tuning parameters and fail to offer a universal solution.  Other works proposed using alternate, equivalent residuals to form the physics loss function. For instance, \citep{patel2022cvpinns,de2022weak_pinns} define residuals using weak entropy conditions and learn weak entropy solutions that capture both shocks and rarefactions \citep{leveque1992numerical}. However, the convergence of these methods has yet to be fully explored and has only seen limited success in practice.

This paper approaches this problem using an operator learning framework, where a parametric operator is trained to map \emph{any input data} to its corresponding LWR PDE solution in a supervised learning setting.  By ``operator learning'', we mean that the method is primarily data-driven compared to the pointwise PDE residual training in PINNs. The Fourier neural operator (FNO) is one such operator that has shown successful results in learning PDE solutions \citep{li2021fno, kovachki2021fno_approx}. To learn solutions of the LWR PDE, we propose a physics-informed variant of the Fourier neural operator ($\pifno$), where an additional physics loss from the integral form of the conservation law is penalized during training. Unlike the differential operator used in \citep{wang2021deeponet_pinns}, the equivalent integral form permits discontinuities and non-differentiabilities in the solutions \citep{patel2022cvpinns,de2022weak_pinns}. The operator is trained offline using data generated from numerical simulations. \emph{Once trained, computing the solution with new inputs,i.e., changing initial/boundary conditions, involves a single forward pass through the trained network.}

The operator learning framework benefits from learning a family of solutions instead of approximating a single solution, as in the PINNs framework. In other words, the PINNs framework requires re-training the neural network for every input data, which is computationally expensive. The operator learning framework alleviates this by offloading the operator optimization to a single offline training stage using a given training dataset.  However, this also has limitations. First, how the trained operators generalize, i.e., how well they perform on previously unseen inputs, is not well understood. Second, there is no guidance on how to choose training data for operator learning, resulting in expensive training procedures to produce operators that perform well \emph{out of sample}. 

To this end, we provide guidance into training data selection for the LWR PDE: solutions generated using input data (i.e., initial and boundary conditions) sampled from random piecewise-constant functions. This specific choice is motivated by the Riemann solutions to the problem \citep{leveque1992numerical} and captures the essential characteristics, namely, shocks and rarefactions. Previous studies have employed similar heuristics, e.g., \citep{wang2021deeponet_pinns} used smooth input data sampled from Gaussian processes to generate data for learning smooth PDE solutions. Our proposed choice of training data distribution is also parametric and has a natural complexity measure, i.e., a measure of data richness. This allows us to train the operator with elementary solutions and then test its generalization to increasingly complex inputs. We also present systematic experiments to study the generalization performance of the trained operator. In experiments, we extensively study the solutions learned for initial value problems (periodic boundary conditions), boundary value problems (Dirichlet boundary conditions), and inverse problems (internal boundary conditions), using the LWR traffic flow model \citep{light1955lwr,richards1956lwr}.

In short, the main contributions of this study (and the remaining sections of the paper) are summarized in the following points:
\vspace{-\topsep}
\begin{itemize}
    \setlength{\parskip}{0pt}
    \setlength\itemsep{0em}
    \item We explore an operator learning framework for learning solutions to macroscopic traffic flow models with arbitrary input data. A unified framework to handle both forward and inverse problems is presented (Section \ref{sec:problem}).
    \item We propose a physics-informed variant of the Fourier neural operator ($\pifno$) for the LWR PDE, where the physics regularizer is derived from its integral form (Section \ref{sec:method}).
    \item We propose a mechanism for selecting training inputs (namely, initial and boundary conditions) 
    that results in superior generalization performance for the trained operator (Section \ref{sec:data}). 
    \item Lastly, we present the numerical results for traffic density predictions for urban signalized roads (Section \ref{sec:res}) and quantify the out-of-sample error performance with respect to the vehicle queue distributions and the number of traffic signal cycles (Section \ref{sec:res2}).
\end{itemize}

\section{Problem formulation}
\label{sec:problem}

\subsection{LWR traffic flow problem}
The LWR model is a continuum description of the flow of vehicles on a road segment. The shock solutions of the LWR represent the sharp changes in traffic conditions, e.g., queue formation and dissipation observed in congested highway traffic and end-of-vehicle queues on urban signalized roads. The LWR model arises as a key component in many real-time traffic state estimation and traffic control algorithms \citep{daganzo1995ctm_net,daganzo2005var,kessel2019tfm}. Below are the details of the model.

Consider a one-dimensional space-time domain $\Omega \subset \mathbb{R} \times \mathbb{R}_{+}$. Denote by $u(x,t): \Omega \rightarrow [0, u_{\max}]$, the density of traffic at position $x \in \mathbb{R}$ and time $t \in \mathbb{R}_+$ and is the average number of vehicles per unit road length or the spatial concentration of vehicles. Let $q(x,t): \Omega \rightarrow [0, q_{\max}]$ be the traffic flux, which is the number of vehicles crossing a road section per unit of time or the temporal concentration of vehicles. The LWR model describes the evolution of traffic density based on the conservation of vehicles:
\begin{equation} 
\label{eqn:lwr_forw}
\begin{split}
& u_t (x,t) + q_x (x,t) = 0, \quad (x,t) \in \Omega, \\
& u(x,0) = \overline{u}_0 (x), \quad (x, 0) \in \Omega_{0}, \\
& u(x,t) = \overline{u}_b (x,t), \quad (x,t) \in \Omega_{b},
\end{split}
\end{equation}
where $u_t \equiv \frac{\partial u}{\partial t}$ and $q_x \equiv \frac{\partial q}{\partial x}$. 
Here $\overline{u}_0$ is the given initial density data defined over the spatial domain $\Omega_{0} \subset \Omega$ and
$\overline{u}_b$ is the given density data defined at the physical road boundaries $\Omega_{b} \subset \Omega$.
For instance, $\overline{u}_b$ corresponds to traffic measurements obtained from point sensors installed at the entry and exit of the road segment. 
We assume that the given boundary condition is well-posed \citep{jabari2016node}.

For the LWR traffic flow model, one typically prescribes a concave flux function $q = f(u)$, $f(u): [0, u_{\max}] \rightarrow [0, q_{\max}]$, which acts as a closure to the above equation \eqref{eqn:lwr_forw}. An example flux function is the Greenshield's fundamental relation \citep{greenshields1935fd}, given by $f(u) = uv_{\rm max} \big(1- u/u_{\rm max} \big)$, where $v_{\rm max}$ is the free-flow speed and $u_{\rm max}$ is the jam density. 
Note $f'(u)$ depends on $u$, which means \eqref{eqn:lwr_forw} is nonlinear for which a simple, shock-free solution does not exist. Thus, we are interested in the weak solution of \eqref{eqn:lwr_forw}, which is composed of shocks and rarefactions \citep{whitham1999waves, leveque1992numerical}. We refer to \eqref{eqn:lwr_forw} as the \emph{forward} problem.

We also consider an associated \emph{inverse} problem,
\begin{equation} \label{eqn:lwr_inve}
\begin{split}
& u_t (x,t) + q_x (x,t) = 0, \quad (x,t) \in \Omega, \\
& u(x,0) = \overline{u}_0 (x), \quad (x, 0) \in \Omega_{0}, \\
& u(x,t) = \overline{u}_p (x,t), \quad (x,t) \in \Omega_{p},
\end{split}
\end{equation}
where instead of the regular boundary data $\overline{u}_{b}$, we are provided with interior boundary data $\overline{u}_{p} (x_p, t_p)$ defined at a set of collocation points within the domain's interior, denoted as $(x_p, t_p) \in \Omega_{p} \subset \Omega$.
For example, $\overline{u}_{p}$ may represent sparse vehicle trajectory measurements obtained from GPS-equipped or connected vehicles, often referred to as probe vehicles. The points $(x_p, t_p) \in \Omega_{p}$ correspond to the spatial and temporal coordinates of the probe vehicle's trajectory within the space-time domain.
However, the input $\overline{u}_{p}$ is a more general specification, encompassing \emph{subsets of} the regular boundary data $\overline{u}_{b}$ mentioned in the forward problem \eqref{eqn:lwr_forw}. The key factor differentiating $\overline{u}_{p}$ from $\overline{u}_{b}$ is that the former does not provide a complete boundary specification, resulting in non-unique solutions to the overall problem. 
To provide a practical example, $\overline{u}_{p}$ can represent probe vehicle trajectory measurements \emph{as well as} measurements collected from a single road boundary. This scenario is relevant to signalized arterials with point sensors at the stopline only.

Due to the incomplete boundary specification, the inverse problem in \eqref{eqn:lwr_inve} is not a well-posed problem and is challenging to solve using conventional numerical solvers such as those based on forward Euler schemes \citep{kessel2019tfm}. Solving \eqref{eqn:lwr_inve} necessitates the use of optimization-based iterative methods to obtain approximate solutions, which largely depends on the nature of input specification $\overline{u}_{p}$. In light of these challenges, the primary objective of this work is to develop a unified framework for learning the solutions to both the forward and inverse problems \eqref{eqn:lwr_forw}-\eqref{eqn:lwr_inve}, regardless of the input specification. Our overarching aim is to make this framework highly accessible for real-world applications and practitioners, thereby easing its practical implementation.

\subsection{Solution using operator learning framework}

\begin{figure*}[!tb]
    \centering
    \includegraphics[width=0.95\textwidth]{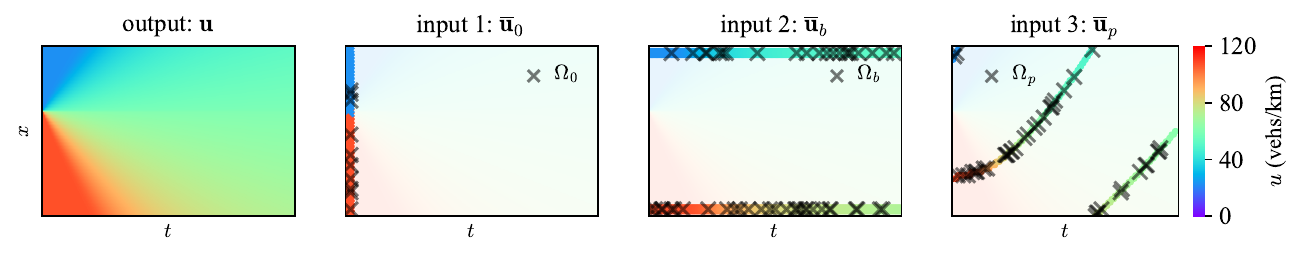} \\
    {\footnotesize (a) Example 1} \\
    \includegraphics[width=0.95\textwidth]{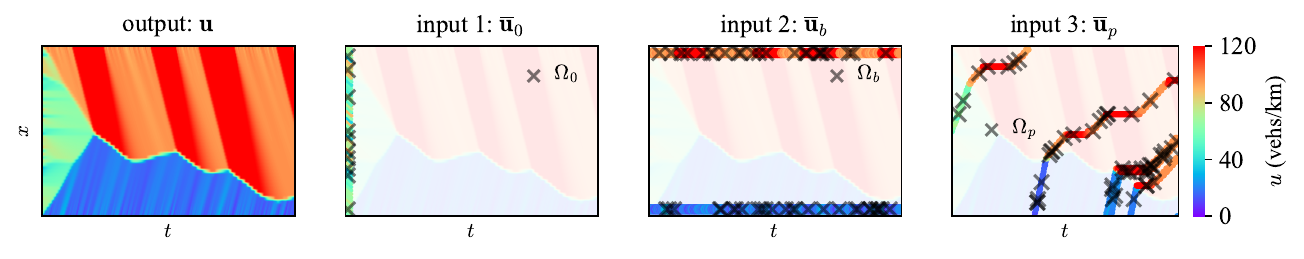} \\
    {\footnotesize (a) Example 2}
    \caption{Three different types of input conditions considered in this study and their representations: initial condition $\widehat{\uu}_{0}$ with domain $\Omega_{0}$, boundary condition $\widehat{\uu}_{b}$ with domain $\Omega_{b}$, and interior condition $\widehat{\uu}_{p}$ with domain $\Omega_{p}$. $\uu$ is the output that defines the solution in the complete domain $\Omega$. Two examples are provided for illustrations.}
    \label{fig:problem_inputs}
\end{figure*}

In this section, we frame the solutions to \eqref{eqn:lwr_forw}-\eqref{eqn:lwr_inve} as an operator learning problem. Denote the complete solution over the domain $\Omega$ as $\uu := \big\{ u(x,t) ~|~ (x,t) \in \Omega \big\}$, arranged into a space-time matrix of dimensions $(m \times n)$, which are the number of space and time discretizations. Similarly, denote the initial data, physical boundary data, and internal boundary data using matrices $\overline{\uu}_0$, $\overline{\uu}_b$, and $\overline{\uu}_p$, respectively, with dimension same as $\uu$. The input conditions ($\overline{\uu}_0$, $\overline{\uu}_b$ and $\overline{\uu}_p$) are defined only at the respective input points in the domain, denoted by $\Omega_{0}$, $\Omega_{b}$, and $\Omega_{p}$ (respectively). The locations where the inputs are unavailable i.e., $(x,t) \in \Omega ~/~ \big\{ \Omega_{0} \cup \Omega_{b} \cup \Omega_{p} \big\}$, are defined as null values, e.g., numerically as $-1$. We illustrate these three input data in Figure \ref{fig:problem_inputs} using two examples.

The original problem \eqref{eqn:lwr_forw}-\eqref{eqn:lwr_inve} is then to reconstruct the complete solution $\uu$ from one or more input condition(s), collectively denoted as $\aa := \big\{ \overline{\uu}_0, \overline{\uu}_b, \overline{\uu}_p \big\}$. We approach this problem with an operator approximation viewpoint, where the goal is to approximate an operator $\GX_{\Theta}$ such that  $\GX_{\Theta}: \aa \mapsto \uu$. This means any input conditions $\aa$ can be mapped to its corresponding solution $\uu$ at the cost of evaluating $\GX_{\Theta}$, which serves as  a surrogate model. The parameters $\Theta$ define the functional class of $\GX_{\Theta}$ (i.e., a set of mappings $\aa \mapsto \uu$). Examples of such operators include Fourier neural operator \citep{li2021fno}, deep-O-net \citep{lu2021deeponet,wang2021deeponet_pinns}, graph kernel operator \citep{li2020multipole}, and wavelet neural operator \citep{tripura2023wno}. 

The operator learning, i.e., finding the optimal parameters $\Theta^{*}$, is performed using an empirical risk minimization approach \citep{murphy2012ml},
\begin{equation}
\label{eqn:problem}
\begin{split}
& \widehat{\uu}_{\rm new} = \GX_{\Theta^*} \big( \aa_{\rm new} \big) , \\
& \Theta^* 
    := \underset{\Theta}{\arg \min} \int_{P} 
    ~\Big[ 
    \LL ~\big( \GX_{\Theta} ( \aa ), ~\uu \big) 
    + 
    \RX ~\big( \GX_{\Theta} ( \aa ) \big) 
    \Big] 
    ~dP \big( \aa, \uu \big) ,
\end{split}
\end{equation}
where $\LL$ is the error in the predicted solution $\mathbf{\widehat{u}} = \GX_{\Theta} \big( \aa \big)$ and true solution $\mathbf{u}$. $\RX$ is a regularizer that incorporates prior knowledge on the operator $\GX_{\Theta}$ or the predicted solution $\mathbf{\widehat{u}}$. The above minimization is over $P (\aa, \uu)$, which is the complete distribution of the input data $\aa$ and its corresponding solution $\uu$. The parameter optimization in \eqref{eqn:problem} is performed only once and done offline. Once optimized, computing solutions for new inputs $\aa_{\rm new}$ involves a single forward pass $\GX_{\Theta^*} \big( \aa_{\rm new} \big)$.

The operator learning framework presented in \eqref{eqn:problem} is generic. There are two elements that primarily dictate its performance: the choice of solution operator $\GX_{\Theta}$ and the data distribution $P \big( \aa, \uu \big)$. Whether there exists such an operator $\GX_{\Theta}$ that can learn a family of the solutions to LWR-PDE is an open research question. Some theoretical studies worked on addressing the approximation errors of neural operators \citep{chen2023operator_cod,ryck2022generic_bounds,kovachki2021fno_approx}, but their results are restricted to well-posed viscous PDEs. In this study, we use the Fourier neural operator as the solution operator $\GX_{\Theta}$ with a slight modification. 
We present this in more detail in Section \ref{sec:method}. Further, the minimization in \eqref{eqn:problem} is over all possible values for $(\aa, \uu)$, i.e., the distribution $P(\aa, \uu)$. However, we only observe a subset $P^{\rm train} (\aa, \uu) \subseteq P(\aa, \uu)$ of the true distribution $P(\aa, \uu)$, in which case the integrand in \eqref{eqn:problem} is replaced by summation. This makes the problem \eqref{eqn:problem} practically feasible. Thus, part of the operator learning problem also involves finding a suitable subset $P^{\rm train} \subseteq P$ so that the operator approximation $G_{\theta^*}$ is valid globally. We present a clever choice of data distribution suitable for learning the weak solutions of LWR PDE in Section \ref{sec:data}.

The framework \eqref{eqn:problem} is also independent of how we choose the input data $\aa$. Depending on the choice of input data (illustrated in Figure \ref{fig:problem_inputs}), we study three different problem settings below. The first two problem settings are forward problems \eqref{eqn:lwr_forw}, and the third is an inverse problem \eqref{eqn:lwr_inve}.
\begin{enumerate}
    \item \textbf{Initial value problem:} The input to these problems is just the initial condition $\mathbf{a}^{\rm ivp} = \big\{ \mathbf{\overline{u}}_0 \big\}$. Such problems arise when the boundary is periodic or free, so explicit solution constraints need not be imposed at the boundary. An example is the simulation of traffic flow on a ring road.
    \item \textbf{Boundary value problem:} The inputs are the initial and Dirichlet-type boundary conditions $\mathbf{a}^{\rm bvp} = \big\{ \mathbf{\overline{u}}_0, \mathbf{\overline{u}}_b \big\}$, a traditional setup for solving LWR PDE. We assume boundary conditions $\mathbf{\overline{u}}_b$ are well-posed, which means the inputs $\mathbf{a}^{\rm bvp}$ can uniquely determine a weak solution for \eqref{eqn:lwr_forw}. These problems arise in simulating traffic flow on an urban road, where the boundary flows are restricted by control measures such as traffic lights. 
    \item \textbf{Inverse problem:} The inputs are the initial and the internal boundary conditions $\mathbf{a}^{\rm ip} = \big\{ \mathbf{\overline{u}}_0, \mathbf{\overline{u}}_p \big\}$. Unlike the forward problems above, this problem is not well-posed due to an unknown boundary condition $\mathbf{\overline{u}}_b$. In other words, $\mathbf{a}^{\rm ip}$ cannot uniquely determine the solution for \eqref{eqn:lwr_inve}. It can have multiple solutions depending on the uncertainty in the boundary conditions, and picking a physical solution would be difficult. These problems arise in practice, where measurements from probe vehicle trajectories $\mathbf{\overline{u}}_p$ are more ubiquitous than measurements at the boundaries $\mathbf{\overline{u}}_b$. It is important to note that $\Omega_{p}$ need not necessarily cover the entire spatiotemporal domain, which can lead to ambiguities in solutions when employing analytical techniques such as the one in \citep{canepa2017network_hjb}. We investigate if an operator learning framework can faithfully reconstruct a physically consistent solution for this type of problem. 
\end{enumerate}

\subsection{Computational aspects of conventional solvers}

\begin{figure}[tb!]
    \centering
    \includegraphics[width=0.65\textwidth]{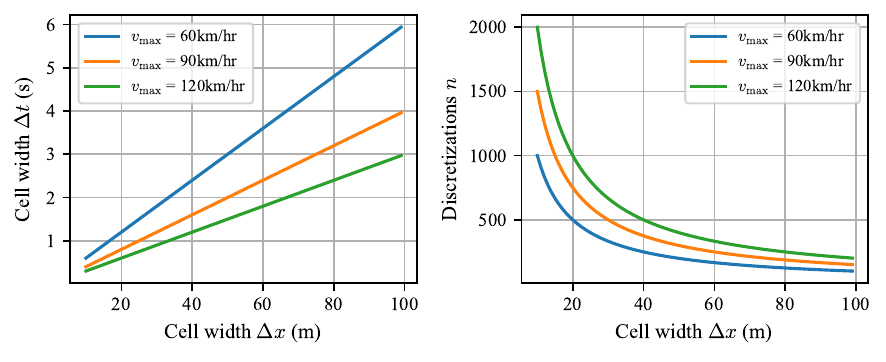}
    \caption{The effect of CFL restriction on the computational mesh size in the context of finite volume numerical schemes.}
    \label{fig:problem_cfl}
\end{figure}

\sloppy We conclude this section with a brief note on the computational aspects of conventional solvers. The forward problem \eqref{eqn:lwr_forw} is generally solved using finite-volume numerical schemes that approximate an average PDE solution in a discrete space-time domain. An example is the Godunov scheme \citep{leveque1992numerical, lebacque1996godunov}. The computations are of the order $O(m n)$ for the one-dimensional LWR PDE, where $m$ and $n$ are the numbers of discretizations in $x$ and $t$ dimensions. However, these methods are primarily limited by choice of discretization size $\Delta t$ and $\Delta x$, which need to satisfy the Courant–Friedrichs–Lewy (CFL) condition \citep{kessel2019tfm}
\begin{equation}
    \Delta t ~\underset{u}{\sup} ~ |f^{'} (u)|  \le \Delta x .
\end{equation}

The CFL condition implies that smaller $\Delta x$ requires a smaller $\Delta t$. Arbitrary discretizations of the domain result in unstable solutions. For instance, if one needs a solution at finer $\Delta x$ but coarser $\Delta t$, it requires first performing computations at finer $\Delta t$ and then aggregating, which mandates \textit{unwarranted} computations. We illustrate this limitation in Figure \ref{fig:problem_cfl}. 
As $\Delta x$ is reduced, the required $\Delta t$ shrinks linearly, but the number of discrete time steps $n$ increases as a power law with exponent $-1$
\begin{equation}
    n \propto \Delta t^{-1} \ge v_{\max}\Delta x^{-1} .
\end{equation}

The inverse problem \eqref{eqn:lwr_inve} has no straightforward solution \citep{tarntola2005inverse_prob}. 
One approach is to consider it as a constrained least squares problem
\begin{equation}
    \label{eqn:inve_1}
    \underset{\uu}{\arg \min} 
    ~~\big\| \mathsf{P}_{\Omega} \big( \uu  - \aa \big) \big\|_{2}^{2} 
    \quad {\rm s.t.} 
    ~~R \big( \uu \big) = 0,
\end{equation}
where $\mathsf{P}_{\Omega}$, ``the binary mask'', projects $\uu  - \aa$ to $\Omega$, i.e., element $i$ of the vector $\mathsf{P}_{\Omega} ( \uu  - \aa )$  is $( \uu  - \aa )_i$ if $i \in \Omega$; it equals 0, otherwise. 
$R(\uu)$ is the prior knowledge of the solution, such as a differential equation, a total-variation bound, or an entropy condition \citep{tarntola2005inverse_prob}. 
If $R$ is linear, \eqref{eqn:inve_1} becomes a quadratic program, which can be solved using convex optimization algorithms. If $R$ is nonlinear, as is the case with many differential equations, one resorts to inexact iterative algorithms \citep{li2022ensemblemc,chuhan2021itsc}.

The physics-informed neural network \citep{raissi2019pinns} is a special case of \eqref{eqn:inve_1}, which solves the following minimization problem for the LWR model \citep{xuan2023physgan_traffic,shi2021pinns_tse,liu2021tse}
\begin{equation}
\label{eqn:inve_2}
\begin{split}
    &\underset{\theta}{\min} ~ 
        \big\|g \big( u_{\theta} \big) \big\|_2^{2} + 
        w_{1} 
        \big\| u_{\theta} \big(x, 0\big) - \overline{u}_{0} \big\|_2^{2} +
        w_{2} 
        \big\| u_{\theta} \big(x_b, t\big) - \overline{u}_{b} \big\|_2^{2}
    \\ 
    &\text{s.t.}~ 
        g\big(u_{\theta}\big) = \partial_{t} u_{\theta} + f'\big(u_{\theta}\big) \partial_{x} u_{\theta} ,
\end{split}
\end{equation}
where a neural network $u_{\theta} : (x,t) \rightarrow u(x,t)$ maps domain points to their corresponding solution and $g(u_{\theta})$ is the physics residual. The problem complexity here is the training convergence of the neural network $u_{\theta}$. In general, the limitation of these approaches to inverse problems \eqref{eqn:inve_1}-\eqref{eqn:inve_2} is that optimization is performed for every input data $\overline{u}_{0}$ and $\overline{u}_{b}$, which is not feasible for online applications.

\section{Methodology I: Solution operator}
\label{sec:method}

\subsection{Fourier neural operator (FNO) as the solution operator}

\begin{figure}[tb!]
    \centering
    \includegraphics[width=0.48\textwidth]{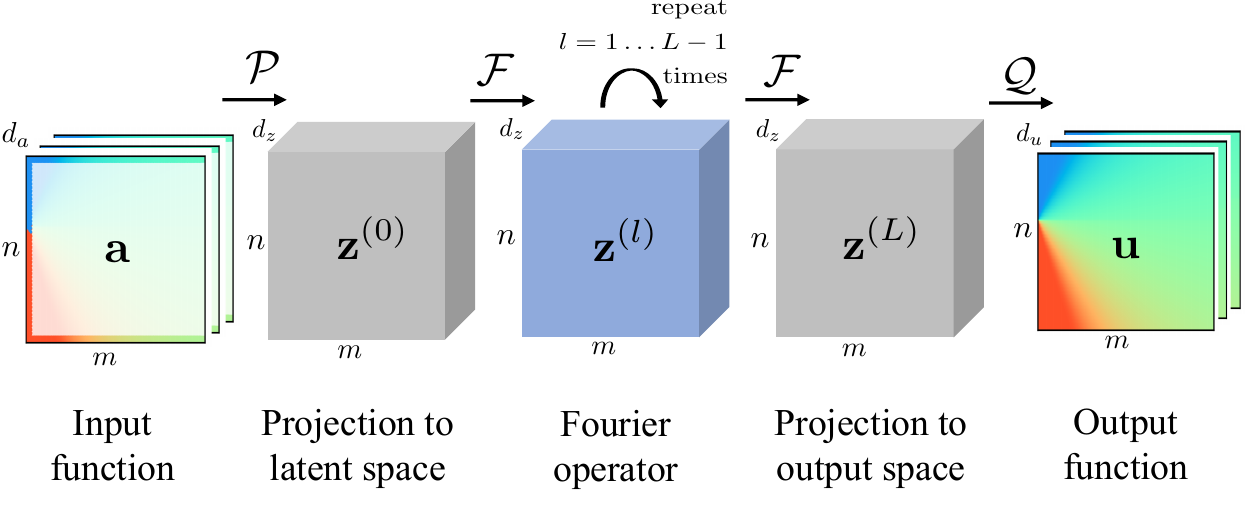} \hfill
    \includegraphics[width=0.48\textwidth]{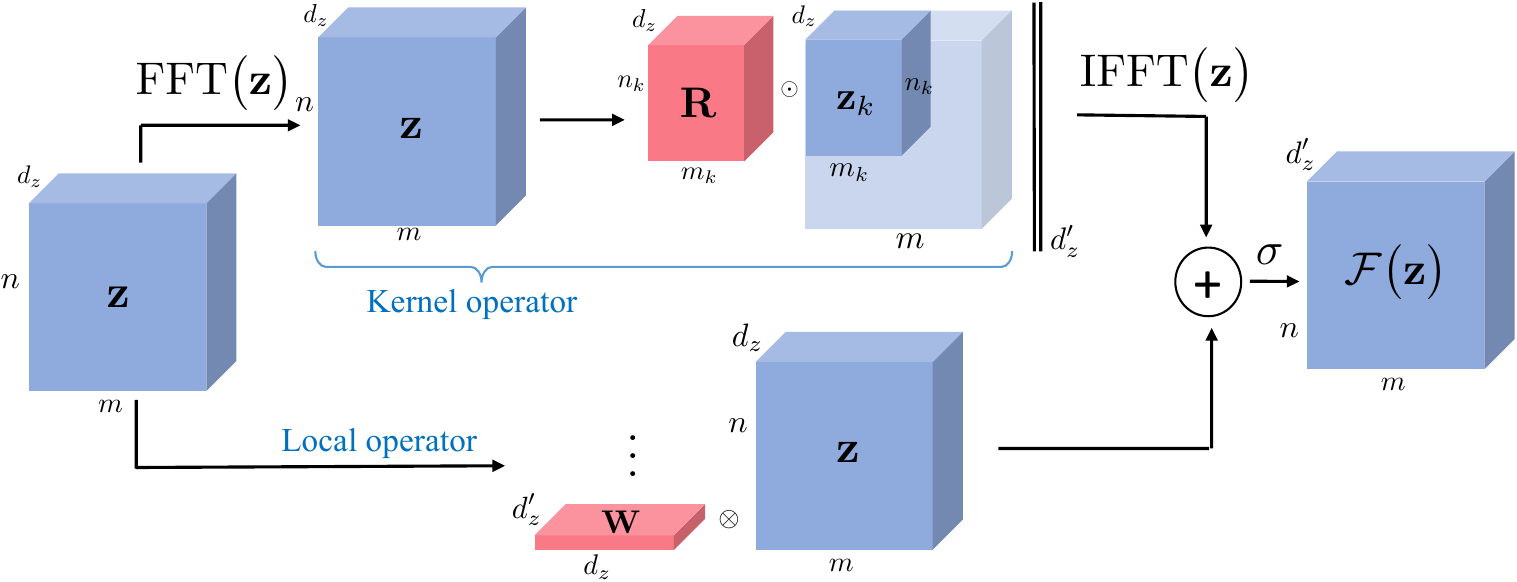} \\
    {\footnotesize (a) Fourier neural operator $\GX_\Theta \big( \aa \big)$ \hspace{1in} (b) Computations in a single Fourier operator $\FF \big( \zz \big)$}
    \caption{Architecture of the Fourier neural operator used as the solution operator in the current paper. (a) Complete model architecture. (b) Single Fourier operator layer. These depictions are adapted and modified from \citep{li2021fno}.}
    \label{fig:fno}
\end{figure}

We use the Fourier neural operator (FNO) model \citep{li2021fno} to approximate the solution operator $\GX_{\Theta}$. We describe the architecture of FNO in this sub-section. Following the standard notations in \citep{li2021fno}, denote the input data $\aa \in \RR^{m} \times \RR^{n}_{+} \times \RR^{d_a}$ and the solution output $\uu \in \RR^{m} \times \RR^{n}_{+} \times \RR^{d_u}$, where $d_a$ and $d_u$ denote the feature dimensions. For the present problem, $d_a = 1$ and $d_u = 1$. The FNO model is defined as a sequence of nonlinear compositions
\begin{equation}
\label{eqn:fno}
\begin{split}
    \mathbf{\widehat{u}} 
        &= \GX_{\Theta} \big( \mathbf{a} \big) 
        &= \Big( \QQ \circ \FF^{(L)} \circ \FF^{(L-1)} \circ \cdots \circ \FF^{(2)} \circ \FF^{(1)} \circ \PP \Big) ~\big( \mathbf{a} \big),
\end{split}
\end{equation}
where $\PP$ and $\QQ$ are projection operators, $\FF^{(l)}$ is a Fourier operator, and $l \in \{1, \dots, L\}$ is the layer index.

The operator $\PP$ lifts the input data $\aa$ to a high-dimensional latent space $\zz^{(0)} = \PP \big( \aa \big), ~~\zz^{(0)} \in \RR^{m} \times \RR^{n} \times \RR^{d_z}$, where $d_z$ is the feature dimension of the latent space. The latent variable $\zz^{(0)}$ gets passed through a series of Fourier operators $\big\{ \zz^{(l+1)} = \FF^{(l)} \big( \zz^{(l)} \big) \big\}_{l=1}^{L}$ until reaching the operator $\QQ$, which project it back to the output space  $\uu = \QQ \big( \zz^{(L)} \big)$. $\PP$ and $\QQ$ are feed forward neural networks with parameters $\Theta_{P}$ and $\Theta_{Q}$. The FNO model is shown in Figure \ref{fig:fno} (a).

Motivated by Green's function for solutions to linear partial differential equations, the original work \citep{li2021fno} defines the Fourier operator $\FF$ as
\begin{equation}
    \label{eqn:fourier}
    \FF (\zz) = \sigma ~\Big( \mathbf{W} \otimes_{1} \zz + {\rm IFFT} \big( \mathbf{R} \otimes_{2} {\rm FFT} (\zz) \big) \Big) ~,
\end{equation}
where ${\rm FFT}$ and ${\rm IFFT}$ are the discrete Fourier transform and the inverse discrete Fourier transform in two dimensions, space and time in this case. $\sigma$ is a nonlinear activation function, acting element-wise. 
The matrix operations $\otimes_{1}$ and $\otimes_{2}$ are defined as
\begin{equation}
\label{eqn:fourier_ops}
\begin{split}
    & \mathbf{W} \otimes_{1} \zz = \Big\|_{n} ~\mathbf{W}^{T} \big[ \zz \big]_{n}
    \quad \quad \text{(local operator)} 
    \\
    & \mathbf{R} \otimes_{2} \Tilde{\zz} = \Big\|_{\Tilde{d_z}} ~ \sum_{d_z} ~\big[ \mathbf{R} \big]_{d_z, \Tilde{d_z}} \odot \big[ \Tilde{\zz} \big]_{d_z, \Tilde{d_z}}
    \quad \quad  \text{(kernel operator)}
\end{split}
\end{equation}
where $\Tilde{\zz} = {\rm FFT} (\zz)$, $\odot$ is the element-wise matrix multiplication, $\|_{i} \cdot$ is the concatenation operator in dimension $i$, and $[~]_{j}$ is the $j^{\rm th}$ component or dimension. 

The first operator (local) in \eqref{eqn:fourier_ops} maps the latent dimension of $\zz$ from $d_z$ to $d'_z$ at all points $(m,n)$ using a matrix $\mathbf{W} \in \mathbb{R}^{d_z \times d'_z}$. This local operation is equivalent to a 1-dimensional convolution. The second operator (kernel) in \eqref{eqn:fourier_ops} performs an element-wise multiplication of $\zz'$ using a tensor $\mathbf{R} \in \mathbb{C}^{m_k \times n_k \times d_z \times d'_z}$, where $\mathbf{R}$ is defined in the complex-space. Figure \ref{fig:fno}(b) visually illustrates these computations. $\mathbf{W}$ and $\mathbf{R}$ are the trainable parameters of a single Fourier operator $\FF$. The central idea of the Fourier operator $\FF$ is the parameterization in the Fourier space. By learning the significant components in the Fourier space, the FNO operator can approximate the solution efficiently. 

The set 
\begin{equation}
    \Theta := \Big\{ \mathbf{W}^{(l)}, ~\mathbf{R}^{(l)} \Big\}_{l=1}^{L} \cup \Big\{ \Theta_{P}, ~\Theta_{Q} \Big\} ,
\end{equation}
is the complete set of the trainable parameters of the solution operator $\GX_{\Theta}$. The operator $\GX_{\Theta}$ is differentiable end-to-end using automatic differentiation. Hence the parameters $\Theta$ can be optimized using any gradient descent-based algorithm. The choice of the loss function and how to incorporate physical knowledge in the optimization is described next.

\subsection{Physics-informed Fourier neural operator $(\pifno)$}

The nature of solutions learned by $\GX_{\Theta}$ depends on the minimization criterion used in the operator learning framework \eqref{eqn:problem}. To enforce physically consistent solutions during training, we choose the prior $\RX$ in \eqref{eqn:problem} as a  physical constraint on the weak solution of LWR PDE. This is discussed below.

\paragraph{\textbf{Physics loss}} 
A straightforward choice for $\RX$ is the point-wise PDE residual, as in conventional physics-informed neural networks \citep{raissi2019pinns}. However, as argued in the previous sections, it does not form a well-defined regularizer due to potential non-differentiabilities or kinks in the solution. Instead, we choose the integral form of the conservation law to form $\RX$. For a compact sub-domain $\Omega^{\rm sub} := \big[x_1, x_2\big] \cup \big[t_1, t_2\big] \subseteq \Omega$, the integral form of a one-dimensional scalar conservation law \citep{leveque1992numerical} is
\begin{equation}
    \label{eqn:integral}
    \int \int_{\Omega^{\rm sub}} ~\big[ u(x,t_2) - u(x, t_1) - \big( q(x_1,t) - q(x_2,t) \big) \big] ~ dx dt = 0 ,
\end{equation}
for all choices of $x_1, x_2, t_1, t_2 \in \Omega$, where $x_1 < x_2$ and $t_1 < t_2$. The integral condition \eqref{eqn:integral} means that the variations in traffic density $u$ over a road section $x\in [x_1, x_2]$ during time $t_1$ to $t_2$ only occur due to a difference in the traffic flux $q$ at the boundaries $x_1$ and $x_2$ within the time interval $[t_1, t_2]$. Condition \eqref{eqn:integral} also holds at shock fronts in the solution.

We enforce condition \eqref{eqn:integral} locally over the discrete domain that is used to define the predicted solution $\widehat{\uu}$. One could also use a different discretization (coarser or finer) to evaluate \eqref{eqn:integral}. The physics prior then becomes
\begin{equation}
\RX \big( \widehat{\uu} \big) 
    = \sum_{(x, t)} 
    \bigg|
    ~\widehat{\uu} \big( x, t+\Delta t \big) - 
    \widehat{\uu} \big( x, t \big) + 
    \frac{\Delta t}{\Delta x} \Big[ 
    \widehat{\qq} \big( x-\Delta x/2, t \big) - 
    \widehat{\qq} \big( x+\Delta x/2, t \big)
    \Big]
    \bigg|
\end{equation}
where $\widehat{\qq}$ is the flux derived from the predicted solution $\widehat{\uu}$. Here, $\Delta x$ and $\Delta t$ denote the size of discrete cells, the choice of which follows the Courant–Friedrichs–Lewy condition \citep{leveque1992numerical}, which for our problem is $\Delta t \le \Delta x / \sup_u |f'(u)|$. The optimal solution operator $\GX_{\Theta}$ should produce predictions $\widehat{\uu}$ that ideally satisfy 
\begin{equation}
    \RX \left( \widehat{\uu} \right) = 0 .
\end{equation}
 
\paragraph{\textbf{Data loss}} We choose the data loss $L$ as the normalized $l_{2}$-norm of the difference in the predicted solution $\widehat{\uu} = \GX_{\Theta} \big( \aa \big)$ and the true solution $\uu$,
\begin{equation}
    \label{eqn:loss_data}
    \LL \left( \widehat{\uu}, \uu \right) = 
    \frac{\big\| \widehat{\uu} - \uu \big\|_2}{ \| \uu \|_2 } .
\end{equation} 
We found minimization using a normalized loss metric to have a stable convergence compared to its unnormalized counterpart. Note that \eqref{eqn:loss_data} naturally incorporates the loss in predicting the given initial and boundary data. 

\paragraph{\textbf{Models}} To evaluate the benefits of physics-informing, we study two different models in this work. Following the operator-learning framework \eqref{eqn:problem}, we consider a vanilla $\fno$ model for which the solution operator $\GX_{\Theta}$ is optimized only using the data loss, 
\begin{equation}
\label{eqn:model_fno}
\underset{\Theta}{\min} ~~ 
    \frac{1}{\big|\mathcal{D}\big|}
    \sum_{i \in \mathcal{D}} ~
    \big\|
    \LL \left( \GX_{\Theta} \big( \aa^{(i)} \big), \uu^{(i)} \right)
    \big\|_{2}^{2}
    \quad \big( \fno {\rm~model} \big) ,
\end{equation}
and a physics-informed $\pifno$ model where the solution operator $\GX_{\Theta}$ is optimized using both the data loss and the physics loss,
\begin{equation}
\label{eqn:model_pifno}
\underset{\Theta}{\min} ~~ 
    \frac{1}{\big|\mathcal{D}\big|}
    \sum_{i \in \mathcal{D}} ~
    \big\|
    \LL \left( \GX_{\Theta} \big( \aa^{(i)} \big), \uu^{(i)} \right)
    \big\|_{2}^{2} 
    +
    \lambda ~
    \big\|
    \RX \left( \GX_{\Theta} \big( \aa^{(i)} \big) \right)
    \big\|_{2}^{2} 
    \quad \big( \pifno {\rm~model} \big) ,
\end{equation}
where $\mathcal{D}$ is the training dataset and $\lambda$ is a relative weight assigned to the physics loss.

We train models \eqref{eqn:model_fno}-\eqref{eqn:model_pifno} for all the problem setups using the same FNO model architecture and training dataset. The hyperparameter $\lambda$ is tuned independently using a brute-force method. The FNO architecture is chosen by trial and error. The model architecture and hyperparameters used in our experiments are detailed in \ref{sec:appendix_fno} and \ref{sec:appendix_lambda}.

\subsection{Computational complexity of $\pifno$ solver}

The computational complexity of the $\pifno$ operator has two parts. First is optimizing the $\pifno$ parameters $\Theta$ for a given training dataset -- the training stage. The second is the forward evaluation of the operator $\GX_{\Theta^*}$ for new inputs -- the inference stage. The computational effort for the training stage depends on the size of $\pifno$ (number of Fourier layers) and training dataset, which we consider less relevant to the present study since training is performed offline with a dedicated computing power. The inference stage or a forward pass is performed in real-time and discussed further below.

The key computational elements of 
a single forward pass of the $\pifno$ are composed of two terms: the global operator and the local operator in \eqref{eqn:fourier}-\eqref{eqn:fourier_ops}. The global operator consists of a two-dimensional Fast Fourier transform followed by an inverse Fast Fourier transform, which has a time complexity of $O (m n (\log m + \log n))$ for uniform spatial and temporal discretizations. 
The local operator is an element-wise matrix multiplication which has a complexity $O(m n)$. So, the total time complexity of the $\pifno$ model is on the order $O(mn (\log m + \log n) + mn)$, which exceeds the Godunov scheme by a factor of $1 + \log (m n)$, when using the same spatio-temporal discretization. However, this complexity measure holds irrespective of the problem type, i.e., it is the same for both the forward and inverse problems, which is a major advantage for $\pifno$. We experimentally compare the computation times later in the paper.

Furthermore, the spectral projection $\zz' = {\rm FFT} (\zz)$ allows one to consider only the significant Fourier components and neglect the insignificant high-frequency components. 
We consider the top $m_k \ll m$ and $n_k \ll n$ components of $\zz$, as shown in Figure \ref{fig:fno}(b) (upper part). Thus, the size of $\mathbf{R}$ is fixed as $(m_k, n_k)$ and independent of the problem size $(m, n)$, which means scalable model complexity. The choice of $(m_k, n_k)$ depends on the regularity of the function $\zz$: low values are preferred for smooth functions, and higher values are preferred for highly-varying functions. Regardless, this factor keeps the model complexity (number of optimization parameters) to be independent of problem size. These advantages are further pronounced for large problem domains. 

\section{Methodology II: Data distribution}
\label{sec:data}

The quality of solutions learned by the $\pifno$ for both seen and unseen inputs partly depends on the choice of the training dataset. Recall the role of data distribution $P^{\rm train}$ in the operator learning framework \eqref{eqn:problem}. Many deep learning-based studies lack a systematic procedure for efficiently choosing the training dataset and evaluating the algorithm's generalization performance. We aim to address these two interrelated aspects in this section.

\subsection{Choice of training data distribution}

The choice of training data depends on the nature of the solution. For instance, \citep{wang2021deeponet_pinns} proposes to use initial conditions sampled from a Gaussian process to generate data for learning the smooth solutions of parabolic PDEs. Following a similar argument and motivated by the Riemann solutions \citep{leveque1992numerical,lebacque1996godunov} used as a base solver for LWR PDE, we propose to use piecewise constant functions as input data $\big( \overline{u}_0, \overline{u}_b \big)$ to generate the solution $\uu$ to train $\pifno$. The piecewise constant input conditions can capture the basic features of the hyperbolic solutions, namely, shocks and rarefactions. However, the nature of the piecewise constant functions depends on the application. In traffic flow models, the initial data $\overline{u}_{0} (x)$ is the initial distribution of vehicles on a road section. We assume vehicles distribute themselves spatially as either one, two, or multiple vehicle queues. This can be modeled using step functions with the number of steps as the number of vehicle queues. For a general setting where all vehicles are unevenly distributed, the number of steps equals the total number of vehicles on the road. Similarly, the boundary condition $\overline{u}_{b} (t)$ specifies the traffic flow at road ends; for e.g., at intersections. On an urban road with traffic lights, we model the effect of red-green vehicles stopping and moving using wavelet-like functions, e.g., Haar wavelets. These wavelet-like functions can capture high-density traffic at the boundary points. Further, boundary data with cycles of non-periodic wavelets can model multiple traffic signal cycles over time.

\begin{figure}[tb!]
    \centering
    \includegraphics[width=0.49\textwidth]{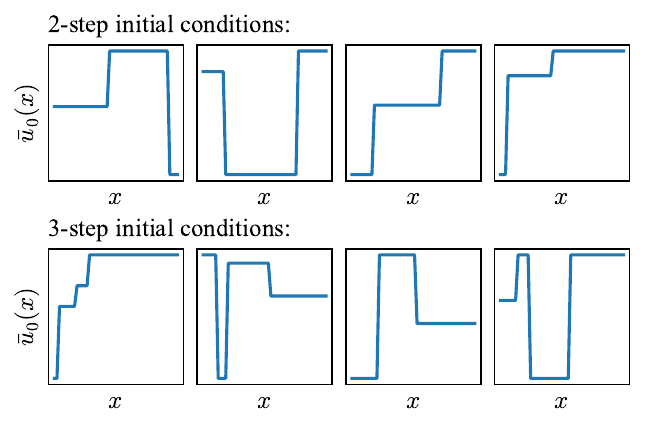}
    \hfill
    \includegraphics[width=0.49\textwidth]{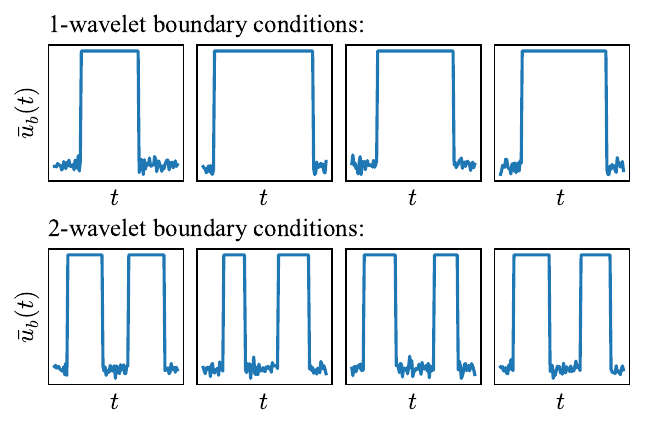} \\
    {\footnotesize (a) Samples of initial conditions \hspace{1.5in} (b) Samples of boundary conditions}
    \caption{Sample paths of input conditions drawn from different families of input distributions. The sample paths are parameterized by the distributions of multi-step functions (initial conditions) and multi-wavelet functions (boundary conditions).}
    \label{fig:data-sample}
\end{figure}

A few examples of the proposed random piecewise constant functions are illustrated in Figure \ref{fig:data-sample}. The width, amplitude, and location of the steps and wavelets used to define initial and boundary data are considered random. We then construct a distribution of such input data, and an input instance is a sample drawn from this distribution. As an example, Figure \ref{fig:data-sample} (a) shows different samples of initial density data with $2$ steps (top row) and $3$ steps (bottom row). Figure \ref{fig:data-sample} (b) shows different samples of boundary density data with $1$ wavelets (top row) and $2$ wavelets (bottom row). This way, one generates a distribution of initial and boundary data, where each distribution is parameterized by the number of steps and wavelets, respectively.

A general procedure to simulate the random multi-step and multi-wavelet functions is described in Algorithms \ref{alg:init_cond} and \ref{alg:bound_cond}. Both algorithms proceed by sequentially adding steps (wavelets) of random widths and heights at random locations in the spatial (temporal) domain. For the boundary data, we partition the temporal domain based on the specified number of wavelets, and each wavelet is randomly located within a single partition. Appropriate bounds are enforced to ensure feasible values. Once a random set of input data are generated, we obtain the corresponding solutions using a numerical solver. We use the Godunov numerical scheme to generate the solutions for the LWR traffic flow model \citep{kessel2019tfm}.

\begin{algorithm}[tb!]
\SetKwInOut{Input}{input}
\SetKwInOut{Initialize}{initialize}
\SetKwInOut{Output}{output}
\SetKwComment{Comment}{/* }{ */}
\caption{Simulate multi-step initial condition}
\label{alg:init_cond}
\Input{number of steps in the initial condition $n^{\rm s} \in \ZZ_{+}$, \\
discretization size in the $x$-dimension $m \in \ZZ_{+}$, \\
solution bounds $u_{\rm min}, u_{\rm max} \in \mathbb{R}_{+}$.}
\Output{sample path $\overline{u}_{0} (x)$ with $n^{\rm s}$ steps}
\Initialize{
    initialize a sample path to a random constant $\overline{u}_{0} (x) = c > 0$, \;
    fixed step height $s_h > 0$, \; 
    maximum step width $s_w = m / n^{\rm s}$, \;
    random integer generator $\varmathbb{U} (\mathrm{start, end})$, \; 
    indices $i = 0, j = 0 .$
}
\For{$\{ 1, \dots, n^{\rm s} \}$ steps}{
    $\,$ randomly locate the next step in the sample path: \\
    $~~~~$ $j \leftarrow \min \big\{\; m-1, ~\varmathbb{U} \big(i, i+ s_w\big) \;\big\}$ \\
    update the sample path value from the new location: \\
    $~~~~$ $\overline{u}_{0}[j:] \leftarrow \overline{u}_{0}[j-1] + ~\varmathbb{U} (-s_h, s_h)$\\
    check for feasibility constraints: \\
    $~~~~$ $\overline{u}_{0} \leftarrow  \max \big( u_{\max}, ~\min \big( u_{\min}, ~\overline{u}_{0} \big) \big)$ \\
    update indices: $i \leftarrow j$ \\
    \Comment*[r]{$\min$ and $\max$ act element-wise}
    }
\end{algorithm}

\begin{algorithm}[tb!]
\SetKwInOut{Input}{input}
\SetKwInOut{Initialize}{initialize}
\SetKwInOut{Output}{output}
\SetKwComment{Comment}{/* }{ */}
\caption{Simulate multi-wavelet boundary condition}
\label{alg:bound_cond}
\Input{number of wavelets in the bound. condition $n^{\rm w} \in \ZZ_{+}$, \\
discretization size in the $t$-dimension $n \in \ZZ_{+}$, \\
solution bounds $u_{\rm min}, u_{\rm max} \in \mathbb{R}_{+}$ .}
\Output{sample path $\overline{u}_{b} (t)$ with $n^{\rm w}$ wavelets}
\Initialize{initialize a sample path using standard Gaussian $\overline{u}_{b} (t) = c + \varmathbb{N}(0, 1, n)$, \;
fixed wavelet width $s_w > 0$, \; maximum partition width $p_w = n / n^{\rm w}$, \;
random integer generator $\varmathbb{U} (\mathrm{start, end})$, \; 
indices $i = 0, j = 0$.}
\For{$k \in \{ 1, \dots, n^{\rm w} \}$ wavelets}{
    $\,$ partition the sample path:
    \\
    $~~~~$ $~\overline{u}_{b}^{\rm par} \leftarrow \overline{u}_{b}[k p_w: (k+1) p_w]$ \\
    randomly locate wavelet position within the partition:
    \\
    $~~~~$ $i \leftarrow \varmathbb{U} (0, p_w/2), \; j \leftarrow \varmathbb{U}(i, p_w)$ 
    \\
    assign maximum solution value for the wavelet location:
    \\
    $~~~~$ $\overline{u}_{b}^{\rm par}[i:j] \leftarrow u_{\max}$ 
    \\
    update the sample path with the updated partition:
    \\
    $~~~~$ $\overline{u}_{b}[k p_w : (k+1) p_w] \leftarrow \overline{u}_{b}^{\rm par}$ \\
    \Comment*[r]{Notation $\overline{u} [i : j]$ denotes slicing the array $\overline{u}$ at indices $i$ and $j$}
    }
\end{algorithm}

\subsection{Evaluation of generalization performance}

\begin{figure*}[t!]
    \centering
    \includegraphics[width=\textwidth]{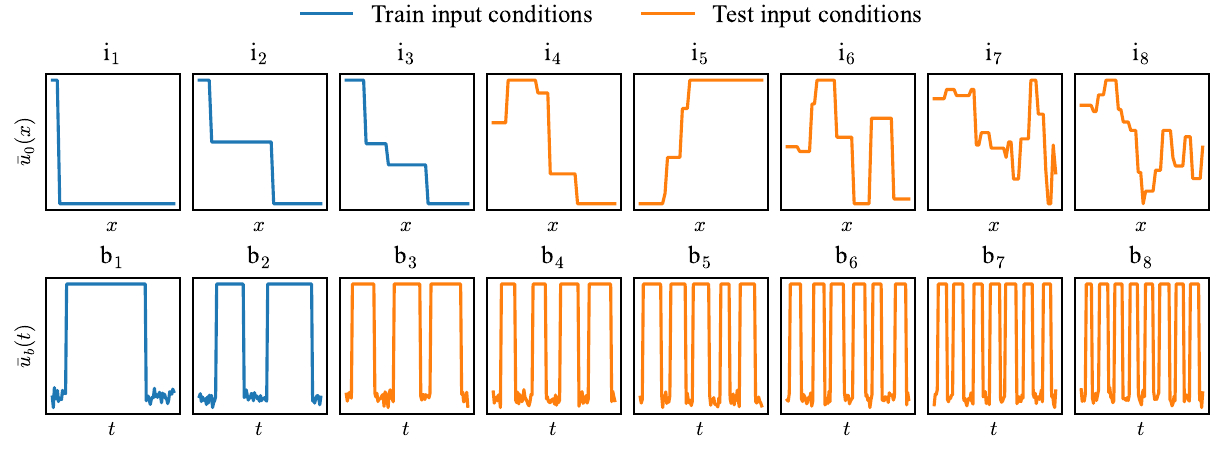}
    \caption{Different input conditions of increasing complexity used in the training data and testing data. The top row shows the initial conditions, and the bottom row shows the boundary conditions. The subscripts denote the number of steps or wavelets.}
    \label{fig:data-inputs}
\end{figure*}

Quantifying the generalization performance is critical for the reliable application of the proposed numerical solver. Generalization performance here refers to the out-of-sample error of the $\pifno$ model on previously unseen input conditions. This is typically measured using $k$-fold cross-validation \citep{murphy2012ml}, which is generally agnostic to the inherent difficulties in the learning problem. For instance, it does not differentiate the difficulty in approximating discontinuous solutions over continuous solutions. We find that it makes more sense to quantify the out-of-sample error as a function of the input conditions of different complexities rather than treating all test samples alike as in $k$-fold cross-validation. The different families of random input functions proposed in the previous sub-section provide a natural grouping for this evaluation.

To this end, we propose a set of systematic experiments to quantify the generalization performance of the $\pifno$ solver. The key is to separate out the training and testing domain as a function of input complexity, which we refer to as the number of steps in the initial data and the number of wavelets in the boundary data. We train the solver using data generated from lower complexity input conditions (simple traffic density dynamics) and test its generalization to input conditions of increasing complexity (complex traffic density dynamics). For instance, the $\pifno$ model is trained with solutions initialized using $1-2$ stepped initial data (homogeneous vehicle queues) and $1-2$ wavelet boundary data (one or two traffic signal cycles). The solver is then tested with solutions of complex dynamics generated from arbitrary initial data (heterogeneous vehicle queues) and multi-wavelet boundary data (multiple red-green traffic signal cycles). We denote a sample path of initial data by i$_\alpha$ and of boundary data by b$_\beta$, where the subscripts $\alpha$ and $\beta$ indicate the number of steps and wavelets (respectively).

Figure \ref{fig:data-inputs} shows sample paths of different initial and boundary data used for training (in blue color) and testing (in orange color). We use initial data of up to three steps (i$_{0}$, i$_{1}$, i$_{2}$ and i$_{3}$) and boundary data of up to two wavelets (b$_{0}$, b$_{1}$ and b$_{2}$) in the training data. The testing data consists of initial data up to forty steps (i$_{4}$ - i$_{40}$) and boundary data up to eight wavelets (b$_{3}$ - b$_{8}$) $-$ the physics of the problem limits these values (for a typical urban road). Note the boundary data shown in Figure \ref{fig:data-inputs} corresponds to the downstream road boundary. For the upstream road boundary, we use boundary data with no wavelets. The goal is to train $\pifno$ with simple solutions and assess the out-of-sample error as input conditions become complex. For the traffic flow example, the input complexity refers to the initial distribution of vehicles on the road ($\overline{u}_0$) and the influence of traffic lights at the road boundaries ($\overline{u}_b$). These two input factors put together can generate complex traffic density dynamics $u$. Figure \ref{fig:data-inputs} illustrates these incremental cases for $\overline{u}_0$ and $\overline{u}_b$. This approach essentially places a limit on the size of training data for operator learning. The results from this systematic evaluation of $\pifno$ are discussed in detail in Section \ref{sec:res2}.

\section{Results I: Numerical experiments}
\label{sec:res}

This section presents a detailed numerical evaluation of the proposed $\pifno$ model for three different problem settings (introduced in Section \ref{sec:problem}): initial value problems (IVP), boundary value problems (BVP), and inverse problems (IP). Throughout the remainder of the paper, the IVP assumes a traffic setting of a closed-loop traffic system (e.g., a ring road), and BVP and IP assume an arterial road section with a free upstream entrance and signal-controlled downstream exit. All road sections have a unit length of $1$ km, and density solutions are predicted for $300$ to $600$ secs. For all road sections, we assume free flow speed $v_{\rm free} = 60$ kmph, jam density $k_{\rm jam} = 120$ vehs/km, and a maximum flow (saturation flow) $q_{\rm max} = 1800$ vehs/hr.

\subsection{Evaluation metrics}

\begin{table}[tb!]
\centering
\small
\caption{Summary of numerical results}
\label{tab:error_metrics}
\resizebox{0.50\textwidth}{!}
{
\begin{tabular}{@{}llllll@{}}
\toprule
\multirow{2}{*}{} & \multirow{2}{*}{Solver} & \multicolumn{2}{l}{Validation error} & \multicolumn{2}{l}{Testing error} \\ \cmidrule(l){3-6} 
 &  & MAE$^{1}$ & Relative $\ell_{2}$$^{2}$ & MAE$^{1}$ & Relative $\ell_{2}$$^{2}$ \\ \midrule
IVP & $\pifno$ & $\mathbf{1.202}$ & $0.025$ & $\mathbf{1.298}$ & $0.033$ \\
 & $\fno$ & $2.260$ & $0.049$ & $2.579$ & $0.070$ \\
 & Diff$^{3}$ & $-46.8 \%$ & \multicolumn{1}{c}{-} & $-49.7 \%$ & \multicolumn{1}{c}{-} \\ \midrule
BVP & $\pifno$ & $\mathbf{1.050}$ & $0.044$ & $\mathbf{1.423}$ & $0.061$ \\
 & $\fno$ & $1.172$ & $0.049$ & $1.596$ & $0.066$ \\
 & Diff$^{3}$ & $-10.4 \%$ & \multicolumn{1}{c}{-} & $-10.8 \%$ & \multicolumn{1}{c}{-} \\ \midrule
IP & $\pifno$ & $\mathbf{2.085}$ & $0.073$ & $\mathbf{2.303}$ & $0.083$ \\
 & $\fno$ & $2.124$ & $0.076$ & $2.333$ & $0.085$ \\
 & Diff$^{3}$ & $-1.8 \%$ & \multicolumn{1}{c}{-} & $-1.3 \%$ & \multicolumn{1}{c}{-} \\ \bottomrule
 \multicolumn{6}{l}{$^{1}$Mean absolute error in vehs/km} \\
 \multicolumn{6}{l}{$^{2}$Relative $l_2$ norm error - unitless metric} \\
 \multicolumn{6}{l}{$^{3}$Percent error reduction of $\pifno$ relative to $\fno$}
\end{tabular}
}
\end{table}

We summarize the solution errors for IVP, BVP, and IP in Table \ref{tab:error_metrics}. The error metric is the average mean absolute error (MAE) of $50$ data samples in units of ${\rm vehs/km}$. Each data sample corresponds to the solution predicted using an initial and/or boundary condition. The MAE metric is in comparison to the reference solution obtained from the numerical scheme. A unitless relative $\ell_2$ norm metric is also shown in the table.
The validation and testing error corresponds to the data samples with input conditions indicated in Figure \ref{fig:data-inputs}. The prediction errors for both $\pifno$ and $\fno$ models are provided for comparison.

We see that the largest MAE incurred for both solvers is $\leq 3.00 ~\mathrm{vehs/km} ~(2.50 \%)$, an acceptable threshold for practical applications. The error metrics for validation and testing samples are similar, implying that the solvers generalize well to unseen input conditions and do not overfit \citep{murphy2012ml}. Further, $\pifno$ incurs a lower average error than $\fno$, indicating the benefit of physics-informed training. We observe this more in the initial valued problems for which the reduction in the prediction error is close to $50\%$. The solvers can also learn solutions for the ill-posed inverse problems though the errors are $2 \times$  higher than those for the forward problems. Also, error metrics for $\pifno$ and $\fno$ only differ slightly ($\approx 2\%$) for the inverse problems.
We will enlighten these results in the following sub-sections. 

\subsection{Initial value problems (IVP)}

\begin{figure}[!t]
    \centering
    \includegraphics[width=0.24\textwidth, valign=t]{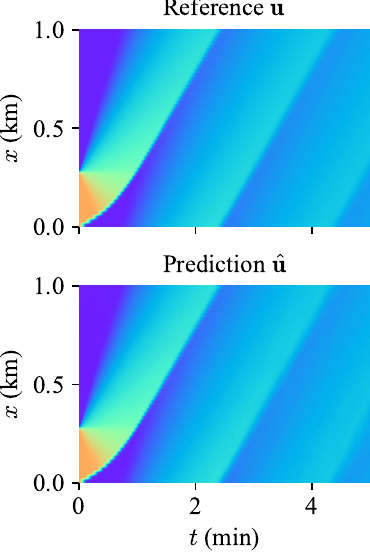} \hfill
    \includegraphics[width=0.24\textwidth, valign=t]{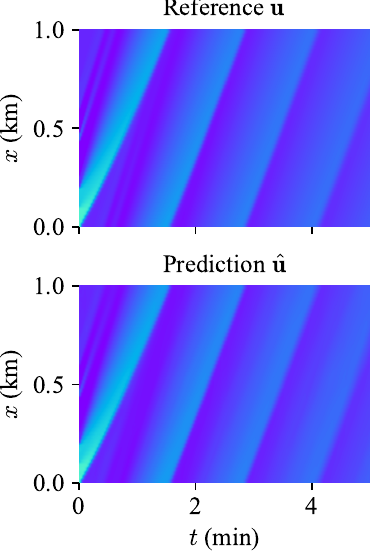} \hfill
    \includegraphics[width=0.24\textwidth, valign=t]{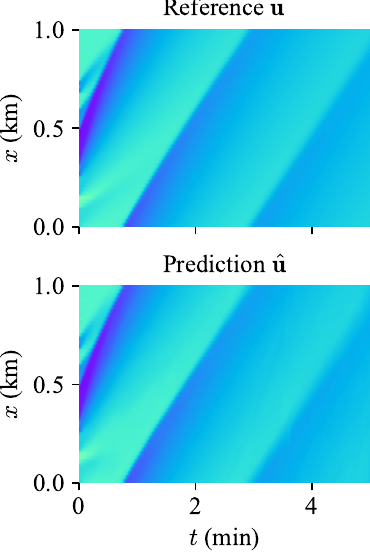} \hfill
    \includegraphics[width=0.24\textwidth, valign=t]{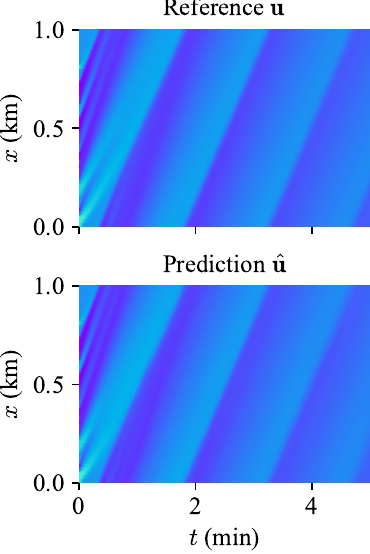} \hfill
    {\footnotesize \hspace{0.0in} (a) Case: i$_{1}$ \hspace{1.0in} (b) Case: i$_{8}$ \hspace{1.0in} (c) Case: i$_{15}$ \hspace{0.90in} (d)  Case: i$_{30}$} \\
    \caption{Sample prediction using $\pifno$ for the initial value problem (bottom row) for different initial data $\overline{u}_{0}$. The reference solution generated from the Godunov scheme is shown (top row) for comparison. Refer to Figure \ref{fig:problem_inputs} for color coding.
    }
    \label{fig:ivp-hmaps}
\end{figure}
\begin{figure}[tb!]
    \centering
    \includegraphics[width=0.49\textwidth]{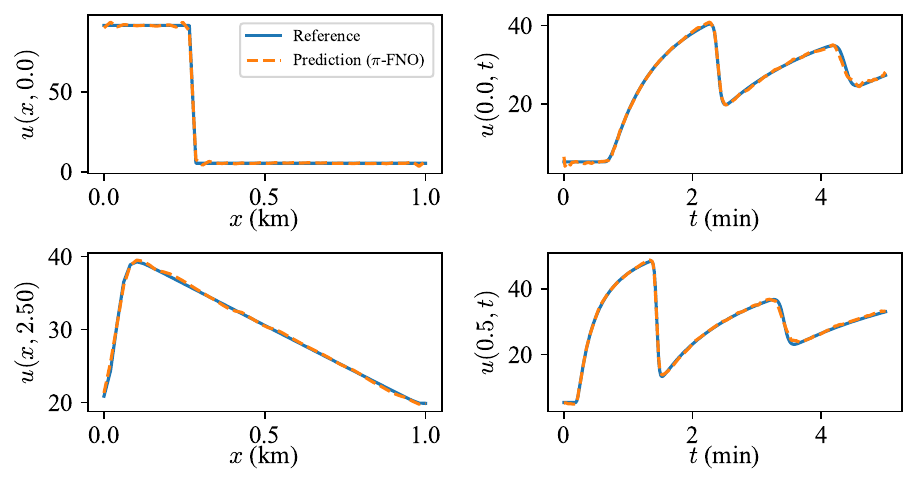} 
    \hfill
    \includegraphics[width=0.49\textwidth]{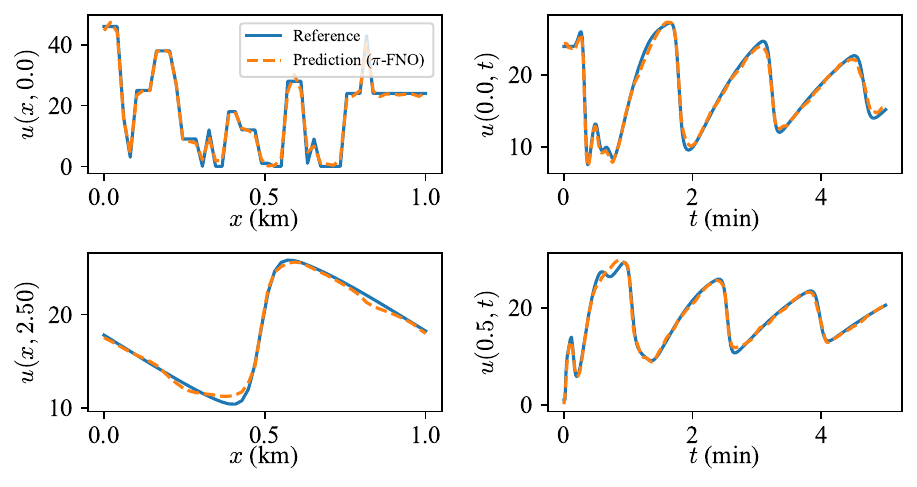} \\
    {\footnotesize (a) Solution profiles corresponding to Figure \ref{fig:ivp-hmaps} (a) \hspace{0.7in} (b) Solution profiles corresponding to Figure \ref{fig:ivp-hmaps} (d) } \\
    \caption{Comparison of the solution profiles at the different $x$ and $t$ values for the initial value problem. The profiles are cross-sections of the heatmaps shown in Figure \ref{fig:ivp-hmaps}.}
    \label{fig:ivp-initialprofile}
\end{figure}

We first present the solutions predicted by $\pifno$ for IVP where only an initial condition $\aa^{\rm ivp}$ dictates the whole solution $\uu$. Four sample results using four different initial conditions $\aa^{\rm ivp}$ are shown in Figure \ref{fig:ivp-hmaps}. The bottom row shows the predicted solution $\widehat{\uu}$ over the complete space-time domain $\Omega$. The top row shows the reference solution $\uu$ obtained from the Godunov numerical scheme for comparison. Figure \ref{fig:ivp-hmaps} (a) is an example from the training data. Figures \ref{fig:ivp-hmaps} (b)-(d) are examples from the testing data. The sub-figure captions denote the respective sample paths used as input data to the solver.

We observe that the predicted solution $\widehat{\uu}$ is in good agreement with the true solution $\uu$ for all the cases. The dissipation and speed of solution discontinuities (shocks), which depend on the initial solution profile, are flawlessly captured in all examples. This suggests a good generalization of $\pifno$ to unseen (out-of-sample) initial conditions. This is evident from test samples in Figures \ref{fig:ivp-hmaps} (b)-(d). We also see that the predicted solutions honor the periodic nature of the boundary conditions. For instance, the solution waves exiting at $x=1$ km emerge exactly at $x=0$ km at the next time-step without any delay or artifacts. Further, we do not explicitly enforce the initial condition $\overline{\uu}_{0}$ (which is input to $\pifno$) in the predicted solution $\widehat{\uu}$. Instead, $\pifno$ learns to predict the initial condition $\overline{\uu}_{0}$. For this specific part of the solution, we see that the solver is performing an identity mapping. To verify these findings, compare the solution profiles at $x=0$ km and $t=0$ min for two examples shown in Figure \ref{fig:ivp-initialprofile}.

From Figure \ref{fig:ivp-initialprofile} (a) and (b) (first row), we observe that the solver accurately replicates the input initial condition and the periodic boundary conditions. The input shock profile $u (x,0)$ and the boundary solution $u (0,t)$ are very accurate. Further, Figure \ref{fig:ivp-initialprofile} (b) is a test sample. Note that previous studies enforce the initial conditions using a bubble function (hard constraint) or using additional weights on the loss function (soft constraint) \citep{patel2022cvpinns}. We found in our experiments that $\pifno$ can sufficiently learn the overall solution dynamics without any additional constraints. A limitation of the $\pifno$ model noted for IVP is that the solutions get dispersed near the end of the time domain, especially for backward-moving waves.

\subsection{Boundary value problems (BVP)}

\begin{figure}[!t]
    \centering
    \includegraphics[width=0.24\textwidth, valign=t]{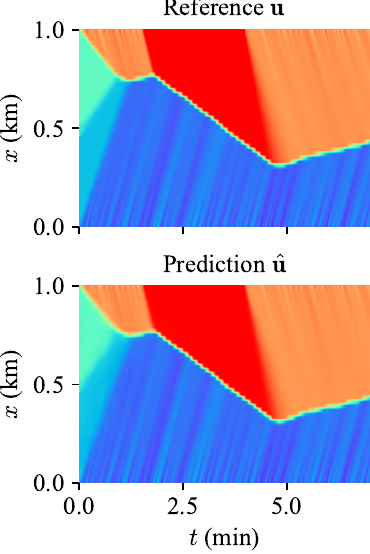} \hfill
    \includegraphics[width=0.24\textwidth, valign=t]{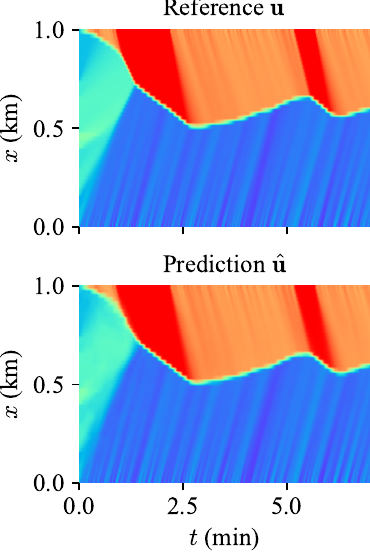} \hfill
    \includegraphics[width=0.24\textwidth, valign=t]{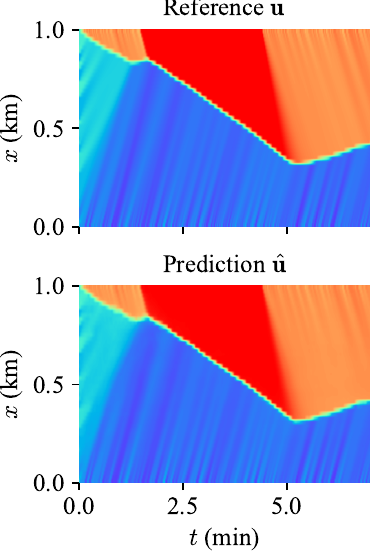} \hfill
    \includegraphics[width=0.24\textwidth, valign=t]{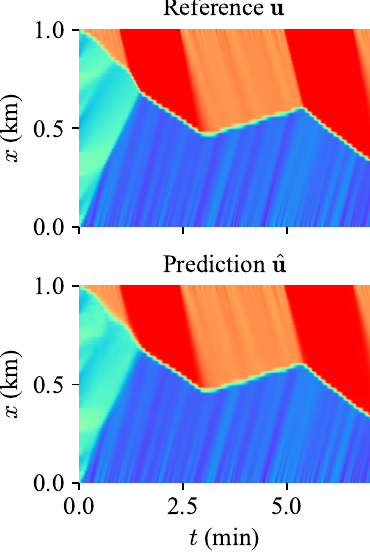} \hfill
    {\footnotesize \hspace{0.00in} (a) Case: (i$_{1}$, b$_{1}$)  \hspace{0.75in} (b) Case: (i$_{10}$, b$_{2}$) \hspace{0.75in} (c) Case: (i$_{20}$, b$_{1}$) \hspace{0.70in} (d)  Case: (i$_{30}$, b$_{2}$)} \\
    \caption{Sample predictions using $\pifno$ for BVP for different initial data $\overline{u}_{0}$. The reference solution generated from the Godunov scheme is shown for comparison. The sub-figures (a) is a training instance, and (b)-(d) are testing instances. Refer to Figure \ref{fig:problem_inputs} for color coding.
    }
    \label{fig:bvp-ic-hmaps}
\end{figure}

\begin{figure}[!t]
    \centering
    \includegraphics[width=0.24\textwidth, valign=t]{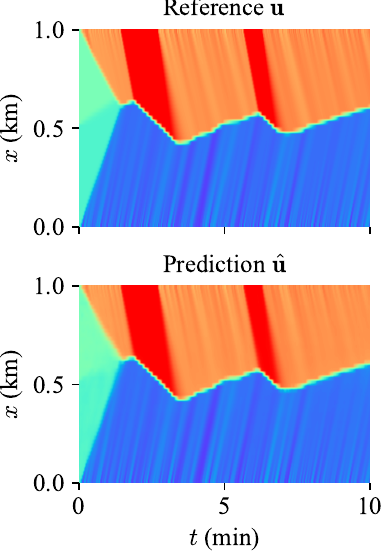} \hfill
    \includegraphics[width=0.24\textwidth, valign=t]{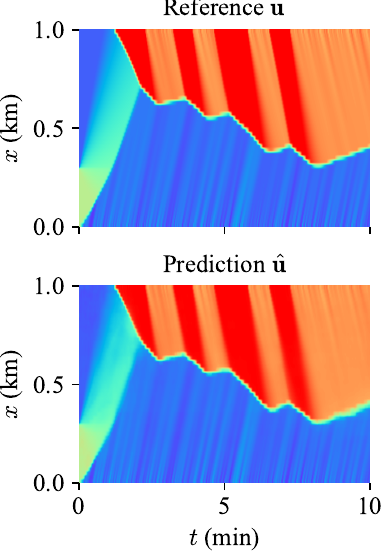} \hfill
    \includegraphics[width=0.24\textwidth, valign=t]{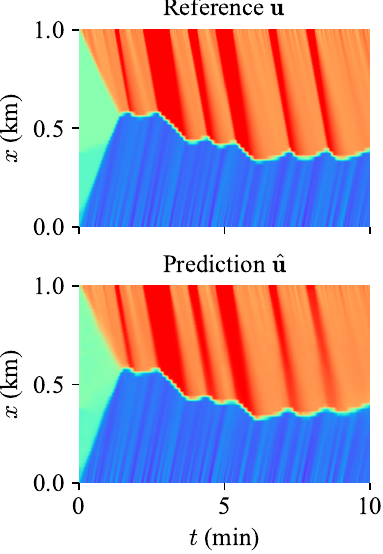} \hfill
    \includegraphics[width=0.24\textwidth, valign=t]{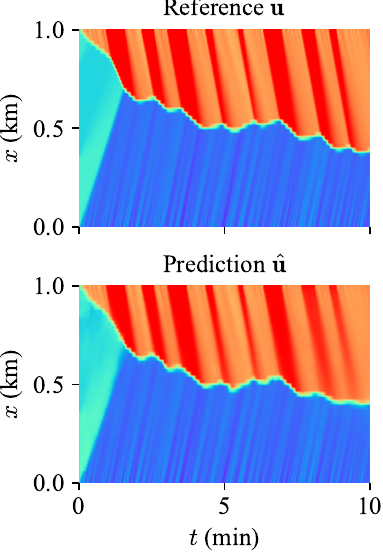} \hfill
    {\footnotesize \hspace{0.00in} (a) Case: (i$_{1}$, b$_{2}$) \hspace{0.75in} (b) Case: (i$_{1}$, b$_{4}$) \hspace{0.75in} (c) Case: (i$_{1}$, b$_{6}$) \hspace{0.70in} (d)  Case: (i$_{0}$, b$_{8}$)} \\
    \caption{Sample predictions using $\pifno$ for BVP for different boundary data $\overline{u}_{b}$. The reference solution generated from the Godunov scheme is  shown for comparison. The sub-figures (a) is a training instance, and (b)-(d) are testing instances. Refer to Figure \ref{fig:problem_inputs} for color coding.
    }
    \label{fig:bvp-bc-hmaps}
\end{figure}

We next present the predicted solutions for BVP where both initial and boundary data $\aa^{\rm bvp}$ collectively influence the density solution $\uu$. Eight example predictions $\widehat{\uu}$ corresponding to eight different input data are shown in Figure \ref{fig:bvp-ic-hmaps} and Figure \ref{fig:bvp-bc-hmaps}. The examples in Figure \ref{fig:bvp-ic-hmaps} are for different initial data $\overline{u}_0$ while the examples in Figure \ref{fig:bvp-bc-hmaps} are for different boundary data $\overline{u}_b$. Figure \ref{fig:bvp-ic-hmaps} (a) and Figure \ref{fig:bvp-bc-hmaps} (a) corresponds to training data. Figures \ref{fig:bvp-ic-hmaps} (b)-(d) and Figures \ref{fig:bvp-bc-hmaps} (b)-(d) are examples from the testing data. The reference solution $\uu$ from the Godunov scheme is also shown for comparison. 

\begin{figure}[!t]
    \centering
    \includegraphics[width=0.49\textwidth]{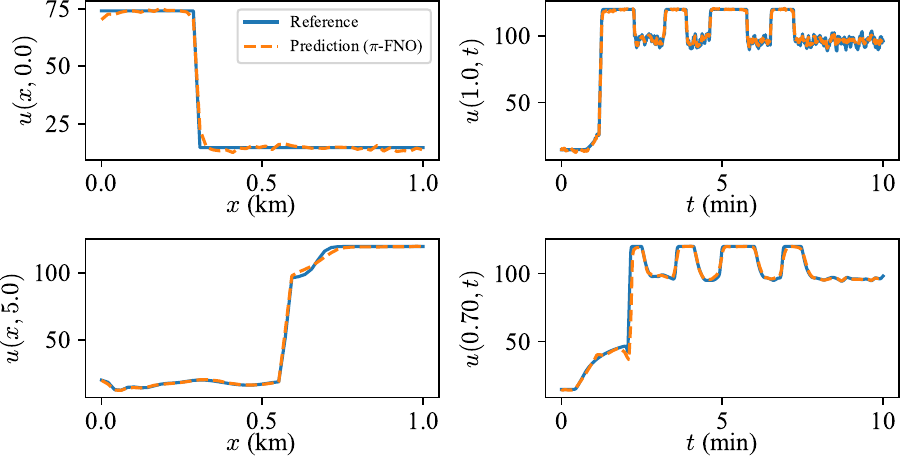} 
    \hfill
    \includegraphics[width=0.49\textwidth]{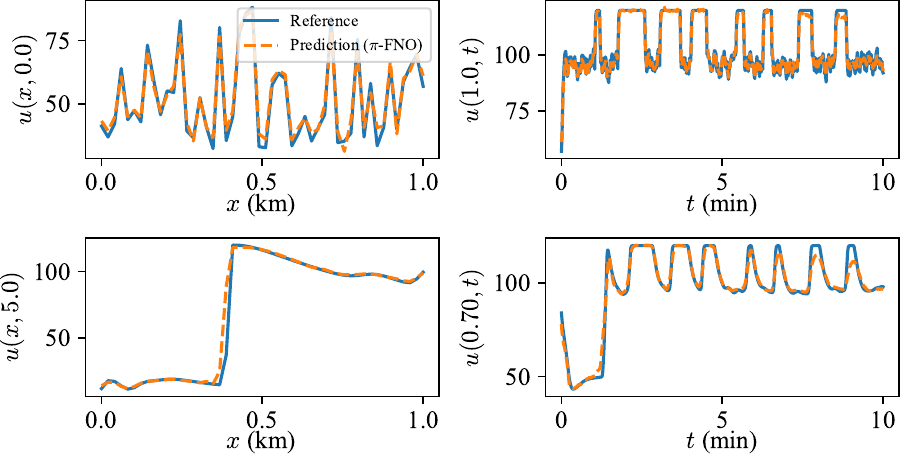} \\
    {\footnotesize (a) Solution profiles corresponding to Figure \ref{fig:bvp-bc-hmaps} (b)  \hspace{0.70in} (b) Solution profiles corresponding to Figure \ref{fig:bvp-bc-hmaps} (d)}  \\
    \caption{Comparison of the solution profiles at the different $x$ and $t$ values for BVP results shown in Figure \ref{fig:bvp-bc-hmaps}.
    }
    \label{fig:bvp-initialprofile}
\end{figure}

We observe that $\pifno$ qualitatively reconstructs the solution for all examples shown in Figures \ref{fig:bvp-ic-hmaps} and \ref{fig:bvp-bc-hmaps}. In particular, the reconstruction of the shockwaves separating different traffic regimes (e.g., free-flow to congestion transition and vice-versa), the rarefaction waves that represent vehicle queues’ dissipation, and the minor waves emanating from the boundaries are all close to the ground truth. The shock waves are mostly sharp, and unphysical artifacts in the predictions are minimal. Similar to IVP, $\pifno$ learns to generalize to more complex initial and boundary conditions. For instance, Figure \ref{fig:bvp-bc-hmaps} (a) is a density solution with two traffic signal cycles (i.e., two wavelets) at the boundary $x=1$ km, which is an instance from the training data. Figure \ref{fig:bvp-bc-hmaps} (b)-(d) are the solution for four, six, and eight traffic signal cycles that are not seen during training. The solver learned the physical queuing dynamics (patterns of the red band) for arbitrary combinations of the number of signal cycles and cycle lengths. This shows the extrapolation capability of the $\pifno$ to boundary conditions of higher complexity. Similar insights on the generalization to different initial conditions can be seen in Figure \ref{fig:bvp-ic-hmaps}.

Similar to IVP, we collectively learn the initial and boundary condition with the complete solution instead of explicitly enforcing them during the model training. To illustrate this, we compare solution profiles at x = $1$ km (upstream boundary condition) and t = $0$ min (initial condition) in Figure \ref{fig:bvp-initialprofile}. The solution profiles correspond to Figure \ref{fig:bvp-bc-hmaps} (b) and Figure \ref{fig:bvp-bc-hmaps} (d), both of which are test samples. The $\pifno$ model perfectly reconstructs the initial data $u(x,0)$ and the major wavelets in the boundary data $u(1,t)$. The profile $u(x,0)$ in Figure \ref{fig:bvp-initialprofile} (b) shows generalization to randomly varying input data. We also noticed for a few samples that the end wavelets are smeared out for boundary data $> {\rm b}_{6}$, as seen in the profile $u(0.7,t)$ in Figure \ref{fig:bvp-initialprofile} (b).

\subsection{Inverse problems (IP)}

\begin{figure}[!t]
    \centering
    \includegraphics[width=0.24\textwidth, valign=t]{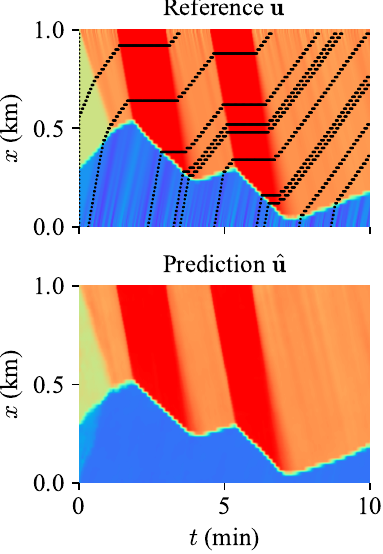} \hfill
    \includegraphics[width=0.24\textwidth, valign=t]{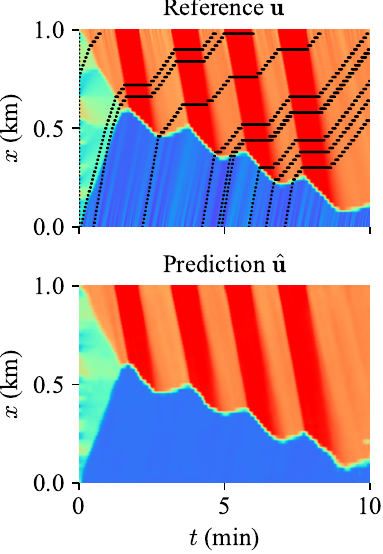} \hfill
    \includegraphics[width=0.24\textwidth, valign=t]{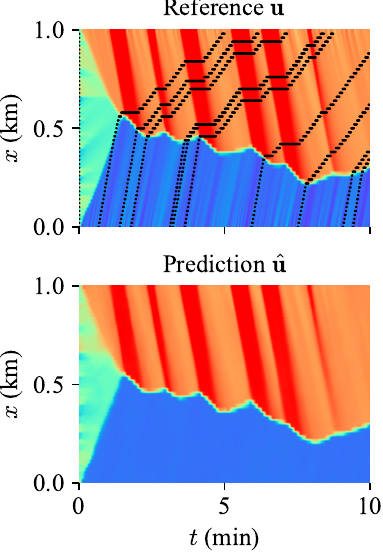} \hfill
    \includegraphics[width=0.24\textwidth, valign=t]{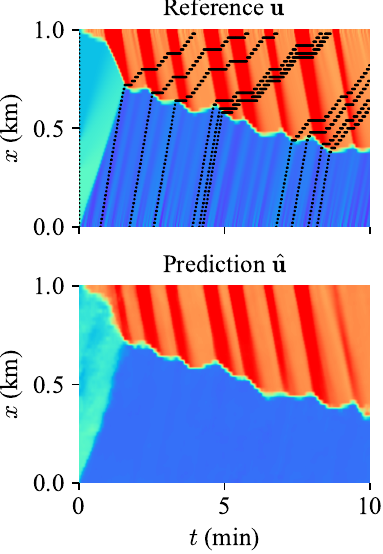} \hfill
    {\footnotesize \hspace{0.0in} (a) Case: (i$_{2}$, b$_{3}$) \hspace{0.75in} (b) Case: (i$_{0}$, b$_{4}$) \hspace{0.75in} (c) Case: (i$_0$, b$_{6}$) \hspace{0.70in} (d)  Case: (i$_{2}$, b$_{7}$)} \\
    \caption{Sample predictions using $\pifno$ for IP for different interior data $\overline{u}_{p}$ (black-dotted curve in the top row). The reference solution generated from the Godunov scheme is shown for comparison. Refer to Figure \ref{fig:problem_inputs} for color coding.}
    \label{fig:bvp-inv_bc-hmaps}
\end{figure}
\begin{figure}[!t]
    \centering
    \includegraphics[width=0.49\textwidth]{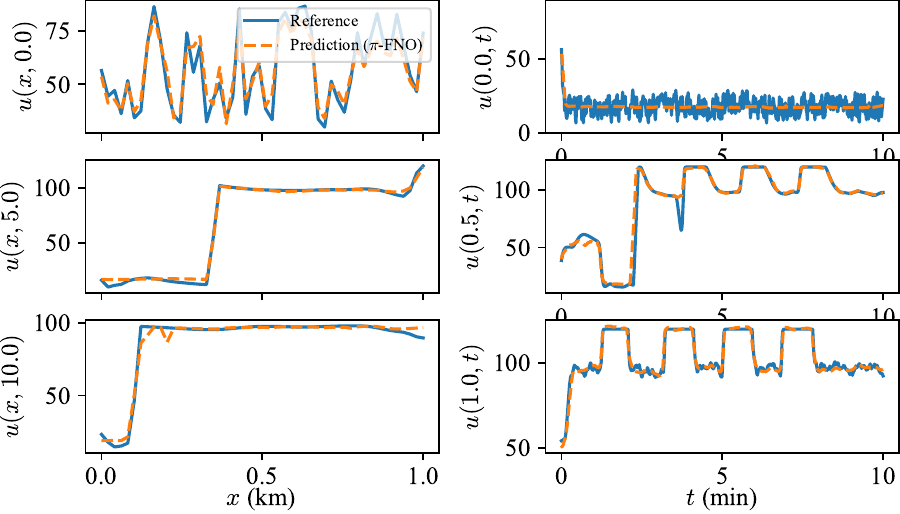} 
    \hfill
    \includegraphics[width=0.49\textwidth]{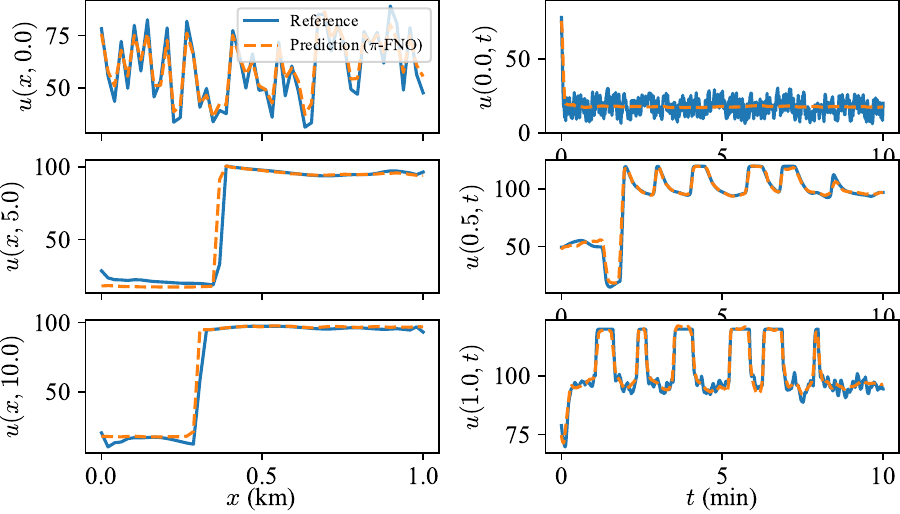} \\
    {\footnotesize (a) Solution profiles corresponding to Figure \ref{fig:bvp-inv_bc-hmaps} (b)  \hspace{0.70in} (b)  (a) Solution profiles corresponding to Figure \ref{fig:bvp-inv_bc-hmaps} (c)} \\
    \caption{Comparison of solution profiles at different $x$ and $t$ values for the inverse problem results shown in Figure \ref{fig:bvp-inv_bc-hmaps}.}
    \label{fig:bvp-inve-profiles}
\end{figure}

We next present $\pifno$ solutions for IP. Figure \ref{fig:bvp-inv_bc-hmaps} shows four sample results corresponding to four different input conditions $\aa^{\rm ip}$. Note that the reference solution $\uu$ is obtained assuming known initial and physical boundary data $\aa = (\overline{\uu}_{0}, \overline{\uu}_{b})$ using the Godunov scheme. In contrast, $\pifno$ only uses the initial data and the internal boundary data $\aa^{\rm ip} = (\overline{\uu}_{0}, \overline{\uu}_{p})$ as input. The location of the internal boundary data (vehicle trajectories) $\Omega_{p}$ is shown as the black dotted curve in the reference solution $\uu$ (top row). Figure \ref{fig:bvp-inv_bc-hmaps} (a) is a training sample, and Figure \ref{fig:bvp-inv_bc-hmaps} (b)-(d) are testing samples. The solution profiles are shown in Figure \ref{fig:bvp-inve-profiles}.

From Figure \ref{fig:bvp-inv_bc-hmaps}, we observe a good agreement between predicted and reference solutions. Despite being ill-posed owing to unknown boundary conditions, $\pifno$ uniquely reconstructs all the observable features in the solution, such as the effect of initial conditions, vehicle queue formation at boundaries, and the discontinuities separating different traffic regimes. The generalizability to higher-order complexity in initial and boundary conditions is in-par with the forward problems. See predictions for three, six, and eight wavelet boundary data in Figure \ref{fig:bvp-inv_bc-hmaps} (b)-(d). Another observation is that the sparse information from the input data $\overline{\uu}_{p}$ can infer physically plausible queueing dynamics. As an example, the last vehicle trajectory in Figure \ref{fig:bvp-inv_bc-hmaps} (b) cannot dictate how the queue dissipates in the future period, i.e., the boundary separating the orange and blue
region. This requires additional input from the upstream boundary ($x=0$ m). But, $\pifno$ inferred this information from the downstream vehicle trajectory and produced a plausible reconstruction. Another instance is the shock dynamics inferred between the $4^{\rm th}$ and $5^{\rm th}$ trajectories in Figure \ref{fig:bvp-inv_bc-hmaps} (d). The distribution of input vehicle trajectories $\Omega_{p}$ will affect the reconstructed solution. Nevertheless, $\pifno$ produces a feasible solution even with unknown boundary conditions.

Finally, it is worth noting that $\pifno$ failed to capture the minor waves emanating from boundaries for IP. Compare the dark blue regions in Figure \ref{fig:bvp-inv_bc-hmaps}. The minor waves are due to small random variations in the boundary condition. This is also evident from the solution profile $u(0.0, t)$ in Figure \ref{fig:bvp-inve-profiles}, where the solver fails to exactly reproduce the randomly fluctuating minor waves. This limitation is expected since the input data $\overline{\uu}_{p}$ does not capture these waves and consequently is not observed in the predicted solution $\widehat{\uu}$. Recall from Figure \ref{fig:bvp-bc-hmaps} that the forward problem uses boundary data as input and $\pifno$ faithfully reconstructs these minor waves. Nevertheless, the predicted solution for the inverse problem captures the primary trend and ignores the minor waves as noise.

\subsection{Benefits of physics-informing}

\begin{figure}[!t]
    \centering
    \includegraphics[width=0.23\textwidth]{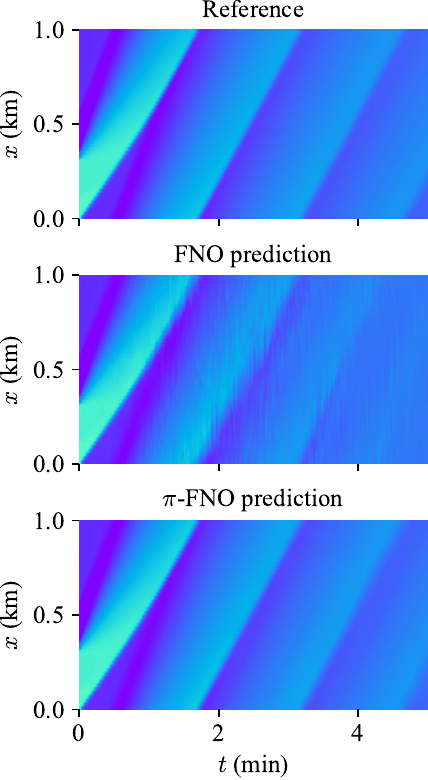}
    $\,$
    \includegraphics[width=0.225\textwidth]{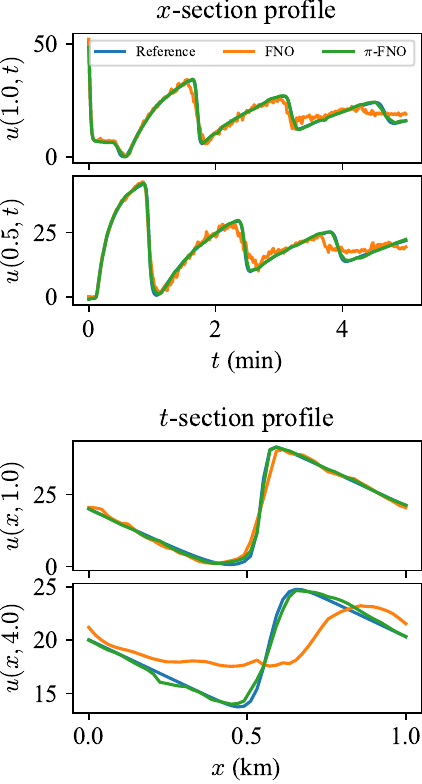}
    \hfill
    \includegraphics[width=0.23\textwidth]{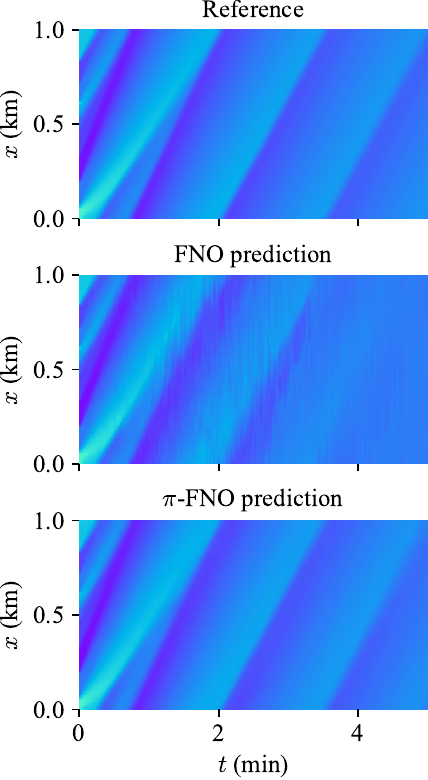}
    $\,$
    \includegraphics[width=0.225\textwidth]{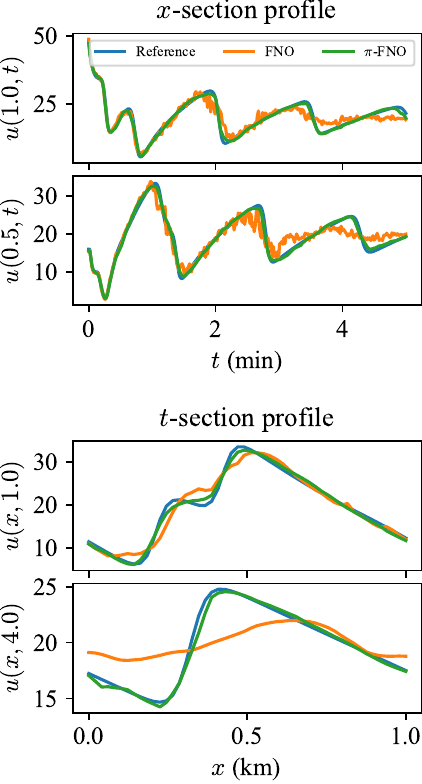} \\
    {\footnotesize (a) Example 1 \hspace{2.5in} (b) Example 2 }\\
    \caption{Comparing solutions predicted using $\pifno$ solver (physics-informed) and $\fno$ solver (physics-uninformed) for IVP. Refer to Figure \ref{fig:problem_inputs} for color coding.}
    \label{fig:ivp-physics}
\end{figure}
\begin{figure}[!t]
    \centering
    \includegraphics[width=0.24\textwidth]{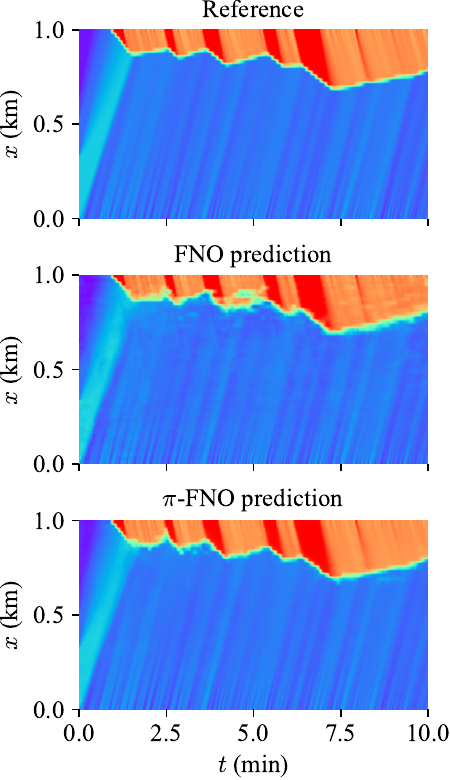}
    \includegraphics[width=0.235\textwidth]{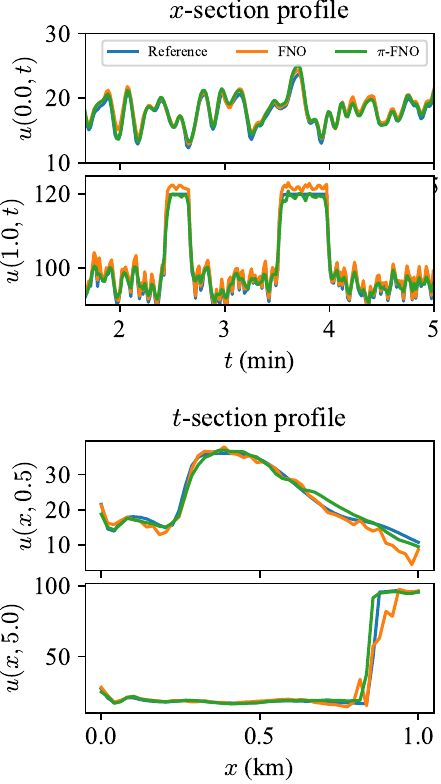}
    \hfill
    \includegraphics[width=0.24\textwidth]{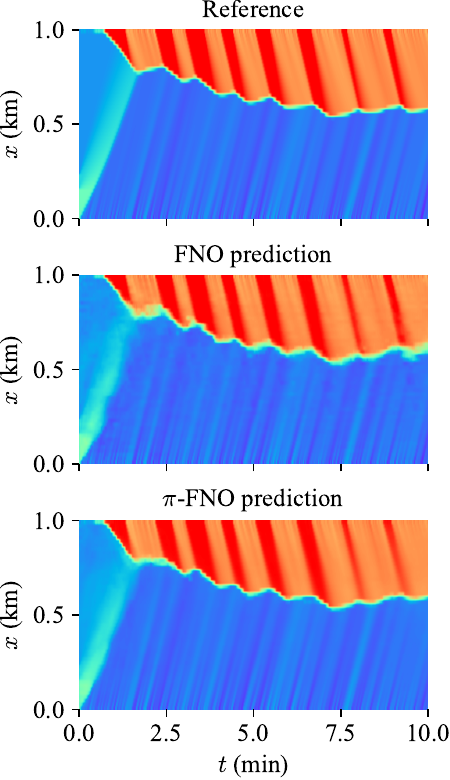}
    \includegraphics[width=0.235\textwidth]{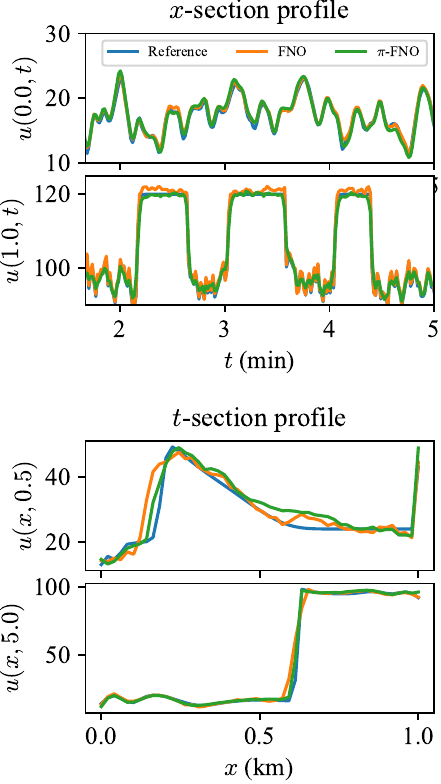} \\
    {\footnotesize (a) Example 1 \hspace{2.5in} (b) Example 2 }\\
    \caption{Comparing solutions predicted using $\pifno$ (physics-informed) and $\fno$ (physics-uninformed) for BVP. Refer to Figure \ref{fig:problem_inputs} for color coding.}
    \label{fig:bvp-physics}
\end{figure}

In this section, we evaluate the benefits of physics-informed training on the predicted solution, especially on the behavior of shocks. Figure \ref{fig:ivp-physics} presents the comparison results for an example IVP, and Figure \ref{fig:bvp-physics} presents that for an example BVP. We recall that the $\fno$ prediction (middle row) is physics-uninformed, and the $\pifno$ prediction (bottom row) is physics-informed. The observed benefits are different for both problems, as discussed below.

From Figure \ref{fig:ivp-physics}, we see that the solution predicted by $\fno$ significantly deviates from the true solution as time progresses. See the $t$-section solution profile $u(x, 4)$ in Figure \ref{fig:ivp-physics}. The profiles indicate that the solution gets smeared to an average value past a certain time, in this case, when $t > 3$ min. Thus, $\fno$ fails to carry over the solution dynamics to longer time periods, possibly due to insufficient inputs. On the other hand, $\pifno$ successfully predicts the complete solution with only a slight deviation from the true solution. This suggests that physics-informed training aids better in learning the long-term dynamics compared to mere data loss. We also found that the FNO solutions are noisy or contain unphysical artifacts that are not interpretable from a physics viewpoint. This occurs even during the initial period $t < 3$ min, as noticed in the $x$-section profiles in Figure \ref{fig:ivp-physics}. This noisy behavior is not observed in the $\pifno$ predictions.

Physics-informing has a different impact on the solutions to BVP, as seen from the comparison results in Figure \ref{fig:bvp-physics}. Both $\fno$ and $\pifno$ reasonably reconstruct all the observable features of the solution. The primary difference appears in the behavior of shocks or solution discontinuities. The heatmaps in Figure \ref{fig:bvp-physics} show that the discontinuities in the $\fno$ solution are smeared-out or noisy. For instance, the blue and red regions in the solution heatmap represent low-density and high-density solutions. The transition from low-density to high-density (thin, light blue region) represents the discontinuous shock solution. The width of this transition region is higher in the $\fno$ solution in comparison to the $\pifno$ solution, which implies $\fno$ predicts a smeared solution instead of a sharp shock. This smeared solution was consistently observed in all the solutions predicted by the $\fno$ model. Further, $\fno$ produces noisy predictions, as evident from the two wavelets (example 1) and three wavelets (example 2) shown in the $x$-section solution profiles $u(1, t)$ in Figure \ref{fig:bvp-physics}. Whereas physics-informing helps minimize these exogenous noises in the predictions and ensure physically consistent solutions.

\subsection{Computation time comparisons}

\begin{table}[tb!]
\centering
\caption{Summary of computational run times}
\label{tab:computational_time}
\resizebox{0.90\textwidth}{!}{%
\begin{tabular}{@{}lllll@{}}
\toprule
\multirow{2}{*}{\begin{tabular}[c]{@{}l@{}} Discretization\\ resolution \\$(m \times n)$\end{tabular}} & \multicolumn{2}{l}{Forward problem} & \multicolumn{2}{l}{Inverse problem} \\ \cmidrule(l){2-5} 
 & \begin{tabular}[c]{@{}l@{}}Conventional method: \\ Godunov scheme \citep{leveque1992numerical}\end{tabular} & \begin{tabular}[c]{@{}l@{}}$\pifno$ model \\ (this paper)\end{tabular} & \begin{tabular}[c]{@{}l@{}}Conventional method:\\ Extended KF \citep{blandind2012tse_mfd}\end{tabular} & \begin{tabular}[c]{@{}l@{}}$\pifno$ model \\ (this paper)\end{tabular} \\ \midrule
$(600 \times 50)$ & $ 0.08 ~(\pm ~0.01) $ & $0.31 ~(\pm ~0.04)$ & $0.55 ~(\pm ~0.07)$ & $0.32 ~(\pm ~0.04)$ \\
$(1200 \times 100)$ & $0.25 ~(\pm ~0.01)$ & $0.75 ~(\pm ~0.07)$ & $3.02 ~(\pm ~0.32)$ & $0.75 ~(\pm ~0.07)$ \\
$(2400 \times 200)$ & $1.16 ~(\pm ~0.13)$ & $4.20 ~(\pm ~0.43)$ & $13.23 ~(\pm ~0.53)$ & $4.20 ~(\pm ~0.43)$ \\ \bottomrule
\end{tabular}%
}
\end{table}

We conclude this section with brief comments on the computational run time of the $\pifno$ model and its comparison with conventional methods. For the conventional methods, we choose the Godunov numerical method \citep{leveque1992numerical} for the forward problem and an extended Kalman Filter implementation \citep{blandind2012tse_mfd} for the inverse problem. The latter is the fastest known scheme for solving inverse problems in traffic flow using the LWR model. It is not the most accurate method (\cite{blandind2012tse_mfd} mention better extensions of the Kalman filter), but our purpose is to compare CPU times. The physics-informed $\pifno$ model is used for both the forward and inverse problems. We report average CPU times for three different domain discretization leves in Table \ref{tab:computational_time}. All methods are evaluated with the same input settings and on the same CPU: an Apple computer with an M1 Mac CPU chipset. From Table \ref{tab:computational_time}, we observe that the Godunov scheme outperforms $\pifno$ in the forward problem, whereas $\pifno$ has the least run time in the inverse problems. This computational run time difference is more pronounced at larger discretizations. 
We add that $\pifno$ run times can be further reduced by efficient parallelization using a GPU chipset. Such parallelization can be easily done with open-source packages such as PyTorch \citep{paszke2019pytorch} with a few additional lines of code. However, how to parallelize the conventional methods has not been addressed in the literature. This is an avenue that we leave to future research.

\section{Results II: Generalization performance}
\label{sec:res2}

\begin{table}[tb!]
\centering
\caption{Summary of generalization error rates}
\label{tab:error_rates}
\resizebox{0.98\textwidth}{!}{%
\begin{tabular}{@{}llll@{}}
\toprule
 & Models & Initial condition & Boundary condition \\ \midrule
Input complexity & \begin{tabular}[c]{@{}l@{}}IVP: FNO\\ IVP: $\pi$-FNO\end{tabular} & \begin{tabular}[c]{@{}l@{}}$0.09 x_{\alpha}^{0.58} + 2.07$\\ $0.08 x_{\alpha}^{0.45} + 1.20$\end{tabular} & $-$ \\ \cmidrule(l){2-4} 
 & \begin{tabular}[c]{@{}l@{}}BVP: FNO\\ BVP: $\pi$-FNO \end{tabular} & \begin{tabular}[c]{@{}l@{}}$0.10 x_{\alpha}^{0.50} + 0.99$\\ $0.08 x_{\alpha}^{0.49} + 0.89$\end{tabular} & \begin{tabular}[c]{@{}l@{}}$\mathbbm{1}_{x_{\beta} \leq 2.50} \big( 1.00 + 0.14 x_{\beta} \big) + \mathbbm{1}_{x_{\beta} > 2.50} \big( 1.00 + 0.14x_{\beta} + 0.06 (x_{\beta}-2.50)^{1.41} \big)$\\ $\mathbbm{1}_{x_{\beta} \leq 2.50} \big( 0.91 + 0.11 x_{\beta} \big) + \mathbbm{1}_{x_{\beta} > 2.50} \big( 0.91 + 0.11x_{\beta} + 0.11 (x_{\beta}-2.50)^{1.16} \big)$\end{tabular} \\ \cmidrule(l){2-4} 
 & \begin{tabular}[c]{@{}l@{}}IP: FNO \\ IP: $\pi$-FNO \end{tabular} & \begin{tabular}[c]{@{}l@{}}$0.01 x_{\alpha}^{0.85} + 1.96$\\ $0.01 x_{\alpha}^{0.91} + 1.96$\end{tabular} & \begin{tabular}[c]{@{}l@{}}$\mathbbm{1}_{x_{\beta} \leq 2.50} \big( 1.92 + 0.08 x_{\beta} \big) + \mathbbm{1}_{x_{\beta} > 2.50} \big( 1.92 + 0.08 x_{\beta} + 0.19 (x_{\beta}-2.50)^{1.07} \big)$\\ $\mathbbm{1}_{x_{\beta} \leq 2.50} \big( 1.95 + 0.04 x_{\beta} \big) + \mathbbm{1}_{x_{\beta} > 2.50} \big( 1.95 + 0.04 x_{\beta} + 0.18 (x_{\beta}-2.50)^{1.17} \big)$\end{tabular} \\ \midrule
Model complexity I$^{\ast}$ & M$_1$ $(m_k=120, n_k=24)$ & $0.08 x_{\alpha}^{0.49} + 0.89$ & $\mathbbm{1}_{x_{\beta} \leq 2.50} \big( 0.92 + 0.11 x_{\beta} \big) + \mathbbm{1}_{x_{\beta} > 2.50} \big( 0.92 + 0.11 x_{\beta} + 0.11 (x_{\beta}-2.50)^{1.16} \big)$ \\
 & M$_2$ $(m_k=90, n_k=20)$ & $0.09 x_{\alpha}^{0.46} + 1.37$ & $\mathbbm{1}_{x_{\beta} \leq 2.50} \big( 1.39 + 0.12 x_{\beta} \big) + \mathbbm{1}_{x_{\beta} > 2.50} \big( 1.39 + 0.12 x_{\beta} + 0.10 (x_{\beta}-2.50)^{1.32} \big)$ \\
 & M$_3$ $(m_k=75, n_k=18)$ & $0.12 x_{\alpha}^{0.46} + 1.81$ & $\mathbbm{1}_{x_{\beta} \leq 2.50} \big( 1.84 + 0.15 x_{\beta} \big) + \mathbbm{1}_{x_{\beta} > 2.50} \big( 1.84 + 0.15 x_{\beta} + 0.12 (x_{\beta}-2.50)^{1.04} \big)$ \\
 & M$_4$ $(m_k=60, n_k=16)$ & $0.16 x_{\alpha}^{0.40} + 2.47$ & $\mathbbm{1}_{x_{\beta} \leq 2.50} \big( 2.52 + 0.16 x_{\beta} \big) + \mathbbm{1}_{x_{\beta} > 2.50} \big( 2.52 + 0.16 x_{\beta} + 0.16 (x_{\beta}-2.50)^{1.14} \big)$ \\ \midrule
Training data I$^{\dagger}$ & D$_1$ $({\rm i}_{0}-{\rm i}_{3}, {\rm b}_{0}-{\rm b}_{2})$ & $0.27 x_{\alpha}^{0.25} + 1.57$ & $\mathbbm{1}_{x_{\beta} \leq 2.50} \big( 1.78 + 0.12 x_{\beta} \big) + \mathbbm{1}_{x_{\beta} > 2.50} \big( 1.78 + 0.12 x_{\beta} + 0.15 (x_{\beta}-2.50)^{1.35} \big)$ \\ & D$_2$ $({\rm i}_{0}-{\rm i}_{6}, {\rm b}_{0}-{\rm b}_{3})$ & $0.26 x_{\alpha}^{0.43} + 0.91$ & $\mathbbm{1}_{x_{\beta} \leq 2.50} \big( 1.19 + 0.08 x_{\beta} \big) + \mathbbm{1}_{x_{\beta} > 3.50} \big( 1.19 + 0.08 x_{\beta} + 0.23 (x_{\beta}-3.50)^{0.97} \big)$ \\
 & D$_3$ $({\rm i}_{0}-{\rm i}_{10}, {\rm b}_{0}-{\rm b}_{4})$ & $0.07 x_{\alpha}^{0.62} + 1.07$ & $\mathbbm{1}_{x_{\beta} \leq 2.50} \big( 1.12 + 0.08 x_{\beta} \big) + \mathbbm{1}_{x_{\beta} > 4.50} \big( 1.12 + 0.08 x_{\beta} + 0.22 (x_{\beta}-4.50)^{0.92} \big)$ \\
 & D$_4$ $({\rm i}_{0}-{\rm i}_{15}, {\rm b}_{0}-{\rm b}_{5})$ & $0.15 x_{\alpha}^{0.39} + 0.85$ & $\mathbbm{1}_{x_{\beta} \leq 2.50} \big( 1.05 + 0.06 x_{\beta} \big) + \mathbbm{1}_{x_{\beta} > 5.50} \big( 1.05 + 0.06 x_{\beta} + 0.47 (x_{\beta}-5.50)^{0.71} \big)$ \\ 
 \bottomrule
 \multicolumn{4}{l}{$^{\ast, \dagger}$ All models are $\pifno$} \\
\end{tabular}%
}
\end{table}

The previous section showed the potential of $\pifno$ in predicting macroscopic traffic states under different input settings. This section extends our evaluation of this deep learning model by examining their generalization performance, by which we mean the prediction error on previously unseen input conditions (out-of-sample error). 
We take an empirical approach to assess the generalization performance, as detailed in the following sub-sections. The results from this analysis are crucial for their reliable application in practice.

\subsection{Error growth as a function of input complexity}

We first study the generalization error rates as a function of input complexity. Recall that input complexity refers to the number of steps in the initial condition (number of different vehicle queues) and the number of wavelets in the boundary condition (number of traffic signal cycles). A higher number of steps and wavelets presumably generate complex solutions, e.g., more interacting shock and rarefaction waves. We train $\pifno$ with solutions consisting of up to \textit{three} steps in the initial data and up to \textit{two} wavelets in the boundary data. We then assess the solution error as a function of the order of input complexity.

\begin{figure}[tb!]
    \centering
    \includegraphics[width=0.35\textwidth]{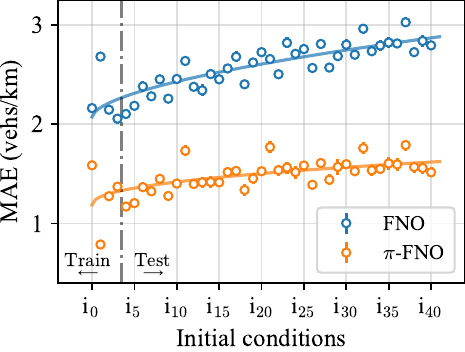}
    \caption{[Input complexity - IVP] Generalization errors for the initial valued problem. The solid curve captures the error growth. The vertical dashed line separates the training and testing domains. }
    \label{fig:ivp-oos}
\end{figure}

The generalization results for IVP are summarized in Figure \ref{fig:ivp-oos}, where we plot the average solution errors for all input conditions i$_{0}$ - i$_{40}$. Each data point is the average mean absolute error (MAE) of $50$ samples. The vertical dashed line separates the training data (i$_{0}$ - i$_{3}$) and testing data (i$_{4}$ - i$_{40}$). The blue data points correspond to $\fno$, and the orange corresponds to $\pifno$. We observe that the out-of-sample error gradually increases as the number of steps in the initial condition increases. The error growth appears sub-linear for both $\pifno$ and $\fno$. 
To quantify the error growth, we fit a generic power law function defined as 
\begin{equation}
y = 
    k_{0} + k_{1} x_{\alpha}^{k_{2}} , 
    \quad 
    \{ k_1 \geq 0, 0 < k_2 \leq 3, k_3 \geq 0 \}, 
    \quad 
    x_{\alpha} \in \big\{ 0, \dots, 40 \big\}
\end{equation}
and is displayed in Figure \ref{fig:ivp-oos} as solid curves. The free parameters $k_{0}$, $k_{1}$, and $k_{2}$ characterize the error growth and are optimized using least squares. 
The exponent $k_{2}$ (predominantly) determines the rate of growth in the power law function and hence the generalization error of the solver. The fitted growth functions are shown in Table \ref{tab:error_rates} (row: Input complexity - IVP).

Table \ref{tab:error_rates} shows that error growth functions for initial value problem are $0.09x_{\alpha}^{0.57}+2.08$ ($\fno$) and $0.08x_{\alpha}^{0.45}+1.20$ ($\pifno$).  We interpret these functions as follows. The exponent value for both solvers is $< 1$, which suggests a \emph{sub-linear} growth in the generalization error with input complexity. An additional unit of step in the initial condition increases the solution error by $+ 0.051x_{\alpha}^{ -0.43} {\rm ~vehs/km}$ for $\fno$ and $+ 0.036x_{\alpha}^{-0.55} {\rm ~vehs/km}$ for $\pifno$. In other words, every additional discontinuity or inhomogeneity in the vehicle queue incurs a \textit{positive} error in the density prediction. All $k_0$, $k_1$, and $k_2$ values for $\pifno$ are lower than $\fno$, suggesting better generalization performance due to physics-informing, in line with our previous results. This is also evident from Figure \ref{fig:ivp-oos}, where the growth function for $\pifno$ is relatively constant compared to that of $\fno$. 

\begin{figure}[!tb]
    \centering
    \includegraphics[width=0.65\textwidth]{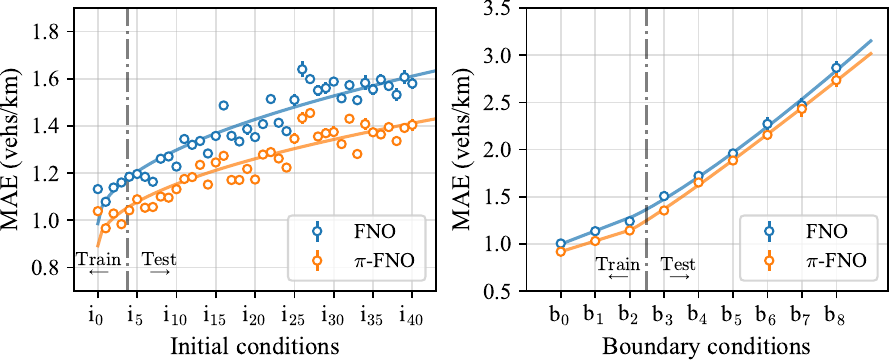}
    \caption{[Input complexity - BVP] Generalization errors for the boundary value problem. The solid curves capture the error growth, and the vertical dashed line separates the training and testing domains. (a) Error curve obtained for different initial conditions and (b) Error curve obtained for different boundary conditions.}
    \label{fig:bvp-oos-for}
\end{figure}

\begin{figure}[!tb]
    \centering\includegraphics[width=0.65\textwidth]{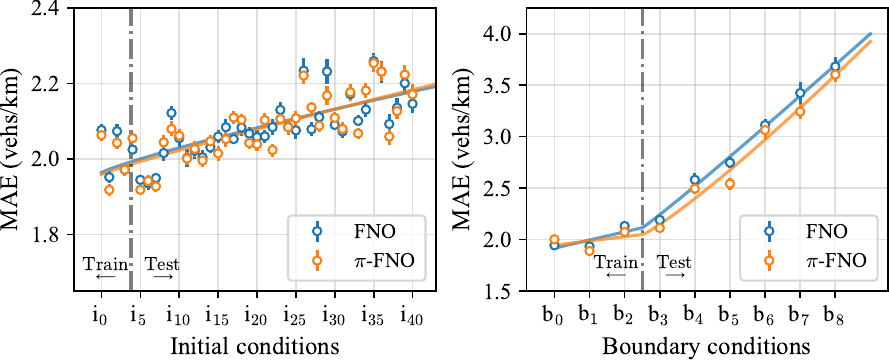} 
    \caption{[Input complexity - IP] Generalization errors for the inverse problem. The solid curves capture the error growth, and the vertical dashed line separates the training and testing domains. (a) Different initial conditions, and (b) Different boundary conditions.}
    \label{fig:bvp-oos-inv}
\end{figure}

Next, we study the error growths for BVP as a function of initial and boundary conditions shown in Figure \ref{fig:bvp-oos-for}. Note that for testing different initial conditions ${\rm i} \in \{ {\rm i}_{0},\dots,{\rm i}_{40} \}$, we use boundary functions from training domain ${\rm b} \in \{ {\rm b}_{0}, {\rm b}_{1},{\rm b}_{2} \}$. And, vice-versa when testing different boundary conditions, i.e., for ${\rm b} \in \{ {\rm b}_{0},\dots,{\rm b}_{8} \}$ we choose ${\rm i} \in \{ {\rm i}_{0}, {\rm i}_{1}, {\rm i}_{2}, {\rm i}_{3} \}$. Figure \ref{fig:bvp-oos-for} (a) shows the error growths for different initial conditions and is similar to that observed for initial value problems in Figure \ref{fig:ivp-oos}. 
Figure \ref{fig:bvp-oos-for} (b) shows the error growth for an increasing number of wavelets in the boundary condition. The vertical dashed line separates the train and test boundary conditions. 

Since there is a clear distinction between train error and test error, we use a piecewise function to model the error growth
\begin{equation}
y = 
    \mathbbm{1}_{x_{\beta} \leq x_{\beta_{0}}} \big( k_{0} + k_{1} x_{\beta} \big) 
    + 
    \mathbbm{1}_{x_{\beta} > x_{\beta_{0}}} \big( k_{0} + k_{1} x_{\beta} + k_{2}(x_{\beta}-x_{\beta_{0}})^{k_{3}} \big), 
    ~~ 
    \{ k_{0} \leq 0, k_{1} \leq 0, k_{2} \leq 0, 0 < k_{3} \leq 3 \}, 
    ~~
    x_{\beta} = \big\{ 0, \dots, 8 \big\}
\end{equation}
where $\mathbbm{1}_{x_{\beta} \leq x_{\beta_{0}}}$ ($\mathbbm{1}_{x_{\beta} > x_{\beta_{0}}}$) is an indicator function which takes the value $1$ if $x_{\beta} \leq x_{\beta_{0}}$ ($x_{\beta} > x_{\beta_{0}}$) and $0$ otherwise. $x_{\beta_{0}}$ is a threshold separating training and testing domain. The first part of the piecewise function (linear) captures training error, and the second part (power law) captures testing error. 
The exponent $k_{3}$ predominantly dictates the rates of error growth and will be used here to interpret the results. The fitted piecewise curves are summarized in Table \ref{tab:error_rates} (row: Input complexity - BVP).

From Figure \ref{fig:bvp-oos-for} (b), we see that the error curve steadily increases with the number of wavelets for both $\pifno$ and $\fno$. This is expected since boundary conditions have a more significant impact on the dynamics of traffic density than the initial condition. The error curve for the training boundary conditions ($\leq$ b$_{2}$) is relatively flat compared to that obtained for testing boundary conditions ($>$ b$_{2}$), which is also expected. Further, Table \ref{tab:error_rates} shows that the exponent value $k_{3}$ for the error curves are $k_3 = 1.41$ ($\fno$) and $k_3 = 1.16$ ($\pifno$), which means a \emph{super-linear} growth in the generalization error, although $\pifno$ incurs a lower exponent value than $\fno$. Further, every additional wavelet (traffic signal cycle) in the boundary data increases the generalization error by $+ 0.08(x_{\beta}-2.50)^{0.41}$ ($\fno$) and $+ 0.02(x_{\beta}-2.50)^{0.16}$ ($\pifno$). However, the latter error rate is nearly constant. Nevertheless, these error growths are relatively faster than those observed for IVP.

Similar insights can be obtained for the IP shown in Figure \ref{fig:bvp-oos-inv} and Table \ref{tab:error_rates} (row: Input complexity - IP). We see a sub-linear error growth for initial conditions and super-linear growth for boundary conditions. However, the shape of error curves is different from that observed for forward problems. For instance, the error growth functions for the initial condition are $0.01x^{0.85}+0.99$ ($\fno$) and $0.01x^{0.91}+1.96$ ($\pifno$), which has higher exponents compared to those obtained for forward problems. 
We also observe a large variation in the errors for the initial conditions in Figure \ref{fig:bvp-oos-inv} (a), which could be the reason for such a higher error rate. On the other hand, the error growth for boundary conditions is similar to the forward problem, except for a higher intercept. 
Also, there is a clear distinction in the error growth for training and testing boundary conditions. Finally, it is also worth noting that $\pifno$ and $\fno$ have a similar error growth for the IP.

\subsection{Effect of model complexity}

\afterpage{%
\begin{figure}[H]
    \centering
    \includegraphics[width=0.65\columnwidth]{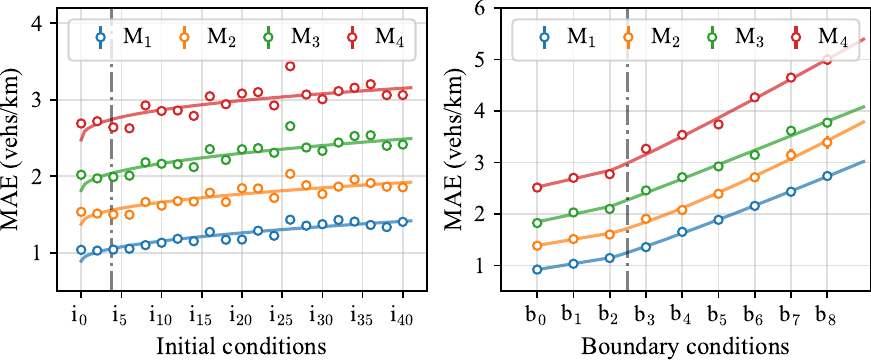}
    \caption{[Model complexity - I] Generalization errors obtained for four $\pifno$ trained different Fourier modes: m$_{1}$ $(m_k=128, n_k=24)$, m$_{2}$ $(m_k=90, n_k=20)$, m$_{3}$ $(m_k=75, n_k=18)$ and m$_{4}$ $(m_k=60, n_k=16)$. The results correspond to the $\pifno$ models for BVP.}
    \label{fig:modelcomplx-oos1}
\end{figure}

\begin{figure}[H]
    \centering
    \includegraphics[width=0.65\columnwidth]{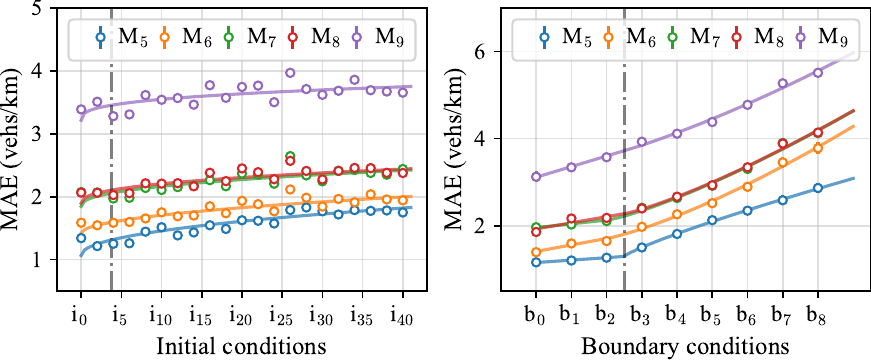} 
    \caption{[Model complexity - II] Generalization errors obtained for five $\pifno$ solvers trained with a different number of Fourier layers: $L = \{ 6, 5, 4, 3, 2 \}$. The results correspond to the $\pifno$ models for BVP.}
    \label{fig:modelcomplx-oos2}
\end{figure}

\begin{figure}[H]
    \centering
    \includegraphics[width=\textwidth]{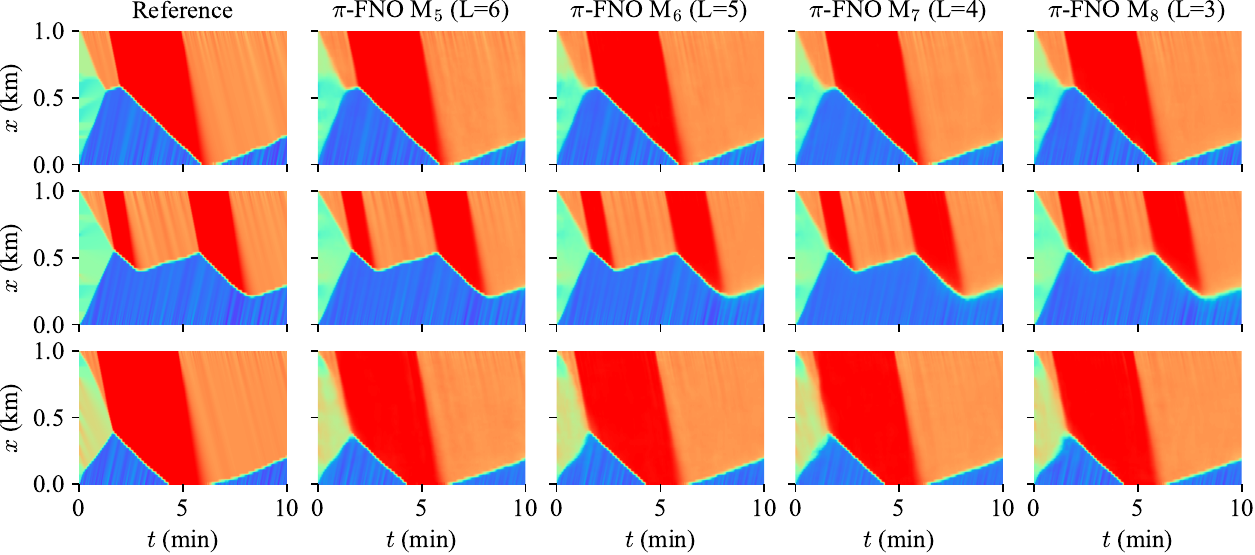}
    \caption{[Model complexity - II] Comparison of solutions predicted using four $\pifno$ solvers trained with a different number of Fourier layers: $L = \{ 6, 5, 4, 3 \}$. The results correspond to the $\pifno$ for BVP. Refer to Figure \ref{fig:problem_inputs} for color coding.}
    \label{fig:modelcomplx-samples}
\end{figure}
\clearpage
}

The previous analysis showed the generalization error curves specific to a chosen $\pifno$ model architecture and training data. We further study how these error curves vary with respect to two key factors discussed in the operator learning framework \eqref{eqn:problem}, namely, the model complexity $\GX_{\Theta}$ (in this sub-section) and training data distribution $P^{\rm ~train}$ (in the next sub-section).

We first study the effect of model complexity on the generalization error of $\pifno$. Model complexity here refers to the learning capacity (approximation power) of the solution operator $\GX_{\theta}$, which we examine using two approaches. 
In the first approach, we change the number of Fourier modes retained in $x$ and $t$ directions ($m_k$ and $n_k$); refer to Figure \ref{fig:fno}.
A higher choice of $m_k$ and $n_k$ corresponds effectively to a higher number of parameters and hence higher model complexity. We consider four levels of model complexity, M$_{1}$ $(m_k=128, n_k=24)$, M$_{2}$ $(m_k=90, n_k=20)$, M$_{3}$ $(m_k=75, n_k=18)$ and M$_{4}$ $(m_k=60, n_k=16)$, and train four different $\pifno$ models with the same training data, keeping every other parameter in the problem fixed. We evaluate the solvers with the same testing data. Figure \ref{fig:modelcomplx-oos1} shows the error curves obtained for all four $\pifno$ models for BVP. We observe that all four models have similar error curves, both for the initial and boundary conditions. This is also evident from the fitted functions in Table \ref{tab:error_rates} (row: Model complexity I). The predominant error term for the initial condition is $\sim x_{\alpha}^{0.49}$ (model M$_{1}$), $\sim x_{\alpha}^{0.46}$ (model M$_{2}$), $\sim x_{\alpha}^{0.46}$ (model M$_{3}$), and $\sim x_{\alpha}^{0.40}$ (model M$_{4}$). The only difference appears in their intercepts which indicate a fitting bias: smaller bias for M$_{1}$ and larger bias for M$_{4}$.

In our second approach to examining model complexity, we change the number of Fourier layers (or depth) $L$ in the solution operator $\GX_{\theta}$. Increasing the depth increases the number of parameters and hence the learning capacity. We train five different $\pifno$ models of different depths, keeping all other parameters fixed: M$_{5}$ $(L=6)$, M$_{6}$ $(L=5)$, M$_{7}$ $(L=4)$, M$_{8}$ $(L=3)$, M$_{9}$ $(L=2)$. The models are trained and tested using the same training and testing data (respectively). Figure \ref{fig:modelcomplx-oos2} shows the error curves, and Table \ref{tab:error_rates_extended} (row: Model complexity II) summarizes the fitted growth functions. Similar to the findings in Figure \ref{fig:modelcomplx-oos1}, models M$_{6}$ to M$_{9}$ have similar error rates but different intercepts. But model M$_{5}$ $(L=6)$ shows a sub-linear error growth $\sim x_{\beta}^{0.80}$ for boundary conditions which shows an improvement. The decreasing intercept with the number of layers indicates an improved bias. We visualize sample predictions using models M$_{5}$ to M$_{8}$ in Figure \ref{fig:modelcomplx-samples}, which shows that shocks are diffused at smaller depths, say $L=\{3,4\}$. These findings imply that model complexity plays a role in better generalization performance, but it comes at the cost of a deeper $\pifno$ model and hence more parameters. 

\subsection{Effect of training data distribution}

\afterpage{%
\begin{figure}[H]
    \centering
    \includegraphics[width=0.65\textwidth]{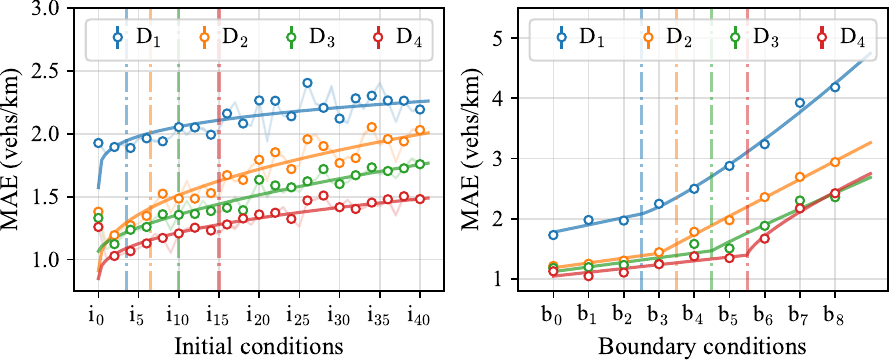}
    \caption{[Training data - I] Generalization errors for four $\pifno$ solvers trained with four different training data distributions: d$_{1}$ $({\rm i}_{0} - {\rm i}_{3}, {\rm b}_{0} - {\rm b}_{2})$, d$_{2}$ $({\rm i}_{0} - {\rm i}_{6}, {\rm b}_{0} - {\rm b}_{3})$, d$_{3}$ $({\rm i}_{0} - {\rm i}_{10}, {\rm b}_{0} - {\rm b}_{4})$ and d$_{4}$ $({\rm i}_{0} - {\rm i}_{15}, {\rm b}_{0} - {\rm b}_{5})$. The dashed lines separate the respective training and testing domains. The results are for BVP. }
    \label{fig:datacomplx-oos1}
\end{figure}

\begin{figure}[H]
    \centering
    \includegraphics[width=0.65\textwidth]{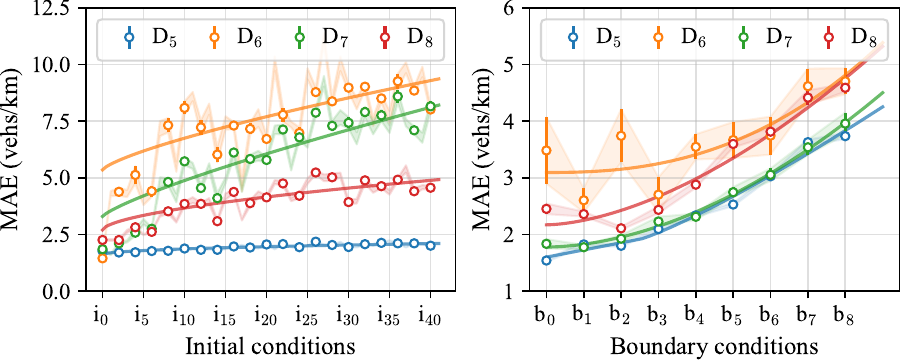} 
    \caption{[Training data - II] Generalization errors obtained for four $\pifno$ solvers trained with four different training data distributions generated using initial conditions: piecewise constant (proposed in this work), random distribution, Gaussian processes, and sinusoid function. The results are for BVP.}
    \label{fig:datacomplx-oos2}
\end{figure}

\begin{figure}[H]
    \centering
    \includegraphics[width=\textwidth]{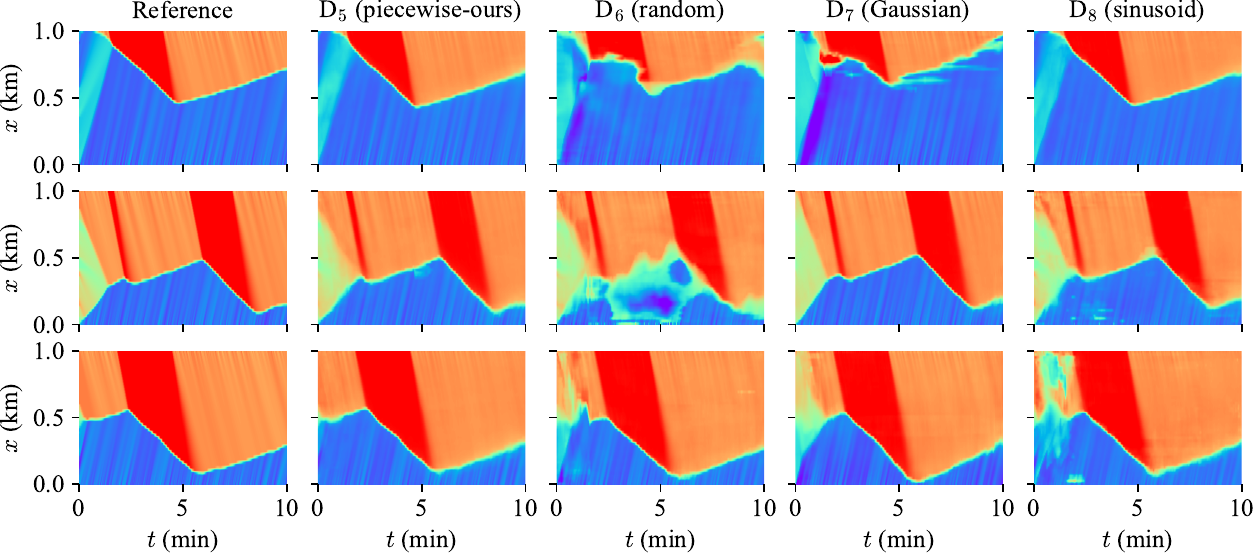}
    \caption{[Training data - II] Comparison of solutions predicted using four $\pifno$ solvers trained with four different training data distributions generated using initial conditions: piecewise constant (proposed in this work), random distribution, Gaussian processes, and sinusoid function. The results are for BVP. Refer to Figure \ref{fig:problem_inputs} for color coding.}
    \label{fig:datacomplx-samples}
\end{figure}
\clearpage
}

We next study the effect of the choice of training data on the generalization performance of $\pifno$. We explore two different methods of choosing training data.
The first method increases the training data richness by augmenting more input conditions (number of steps and wavelets) in the training data.
We consider four increasing levels of data richness: D$_{1}$ $({\rm i}_{0} - {\rm i}_{3}, {\rm b}_{0} - {\rm b}_{2})$, D$_{2}$ $({\rm i}_{0} - {\rm i}_{6}, {\rm b}_{0} - {\rm b}_{3})$, D$_{3}$ $({\rm i}_{0} - {\rm i}_{10}, {\rm b}_{0} - {\rm b}_{4})$ and D$_{4}$ $({\rm i}_{0} - {\rm i}_{15}, {\rm b}_{0} - {\rm b}_{5})$. The data samples are generated as explained in Section \ref{sec:data}. We train four $\pifno$ models using the four training datasets D$_{1}$, D$_{2}$, D$_{3}$ and D$_{4}$, and evaluate them with the same testing data. The results are summarized in Figure \ref{fig:datacomplx-oos1} and Table \ref{tab:error_rates} (row: Training data I). The four vertical dashed lines in Figure \ref{fig:datacomplx-oos1} separate the training and testing domain of the four solvers.
We observe that 
all four error curves have different shapes, especially with respect to the boundary conditions shown in Figure  \ref{fig:datacomplx-oos1} (b). The predominant growth factor (boundary condition) is $\sim x_{\beta}^{1.35}$ (model D$_{1}$), $\sim x_{\beta}^{0.97}$ (model D$_{2}$), $\sim x_{\beta}^{0.92}$ (model D$_{3}$) and $\sim x_{\beta}^{0.71}$ (model D$_{4}$), which shows a decreasing trend. In other words, the error curves shifted from a \textit{super-linear} growth to \textit{sub-linear} growth as we increased the richness of training data. 
The error curves also shifted down marginally, implying a lower bias. 

In the second method, we choose input data other than random piecewise constant functions (proposed in Section \ref{sec:data}) for generating the training data. We study how the generalization performance changes as we change the whole family of training data $P^{\rm train}$. For this, we generate data using initial conditions sampled from three random families: a uniform distribution, a Gaussian process \citep{wang2021deeponet_pinns}, and a sinusoid function\footnote{The sinusoid wave function used is $\overline{u}_{0} = a + b \sin (w x)$, where $a$, $b$ and $w$ are random parameters.}. We train three respective $\pifno$ models and compare their performance to $\pifno$ trained using the proposed piecewise constant data: D$_{5}$ (piecewise constant - ours), D$_{6}$ (random), D$_{7}$ (Gaussian), and D$_{8}$ (sinusoid). Figure \ref{fig:datacomplx-oos2} shows the error curves, and Table \ref{tab:error_rates_extended} summarizes the fitted growth functions. 
We see that all error curves have different shapes, both for initial and boundary conditions. 
Clearly, $\pifno$ trained using data sampled from piecewise constant functions performs better, and those trained with random and Gaussian distributions perform the worst.  
The predominant growth factor  (initial condition) is $\sim x_{\alpha}^{0.45}$ (ours), $\sim x_{\alpha}^{0.75}$ (random), $\sim x_{\alpha}^{0.80}$ (Gaussian), and $\sim x_{\alpha}^{0.59}$ (sinusoid), which indicates that poor training data can even lead to near-linear error growth. We also find a significant error variation for the three solvers (shaded curves in Figure \ref{fig:datacomplx-oos2}). These observations are also in parallel with the density solutions compared in Figure \ref{fig:datacomplx-samples}. Thus,  piecewise constant input data proved effective in learning the discontinuous solutions of the LWR PDE model. These results also emphasize the importance of training data distribution and how a poor choice, such as uniform random distribution, can result in a poorly generalizing solution operator.

\section{Results III: Generalization to real-world traffic data}

Following our previous analyses using simulated data, we conclude our assessment of the $\pifno$ model by evaluating its generalization to real-world traffic data. We test the $\pifno$ model using an open-source real-world traffic dataset called pNeuma, which is a high-resolution vehicle trajectory dataset collected from a busy traffic network in Athens, Greece \citep{barmpounakis2020pneuma}. The traffic network primarily consists of one-way streets with signalized and unsignalized intersections. The trajectory records include position coordinates, instantaneous speeds, accelerations, and vehicle types. This open-sourced requires pre-processing as detailed in \citep{barmpounakis2020pneuma,barmpounakis2020pneuma_lanedet}. However, the pNeuma dataset has a different traffic composition, network layout, and traffic
density distribution than the numerical simulation data used for training the $\pifno$ model, which might lead to sub-optimal prediction results when tested directly.

\begin{figure}[!t]
    \centering
    \includegraphics[width=\textwidth]{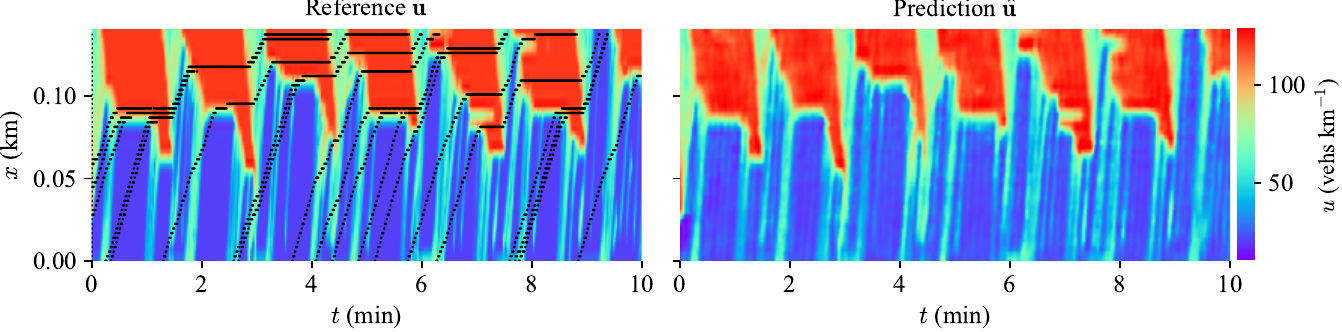}
    \caption{Comparison of reference and $\pifno$ model predicted density field for an intersection link in the pNeuma dataset.}
    \label{fig:pneuma_sam1}
\end{figure}

To address this distribution shift, we perform an additional fine-tuning step in which the trained $\pifno$ model parameters are updated with newer data samples from the pNeuma dataset as a post-training step. This fine-tuning method is a popular transfer learning technique studied in the literature, which broadly deals with methods for transferring deep learning models trained using data from a source domain to a target domain, where the source and target domains are related \citep{yu2019transfer_tse,shai2003multitask_learning}. In our specific case, the source domain is simulation, and the target domain is the real world. We fine-tune the $\pifno$ model with $25$ validation data samples from the pNeuma dataset for $10$ training epochs with a lower learning rate of $10^{-4}$. Compared to the original training, which uses $12,000+$ training samples and $500$ training epochs, this fine-tuning step is notably inexpensive. A sample prediction of the fine-tuned $\pifno$ model from the pNeuma dataset is shown in Figure \ref{fig:pneuma_sam1}. The inputs to the problem are the initial densities and trajectories obtained from probe vehicles, imitating the inverse problem setting studied earlier in the paper. The test sample corresponds to a traffic snapshot observed from an arterial road section controlled by a downstream traffic signal. The length of the arterial road is approximately $140$ m and the estimation period is $10$ min. The maximum traffic density observed in the road section is $120$ veh/km.

As observed in Figure \ref{fig:pneuma_sam1}, the reference and predicted solution are in close agreement, concluding the applicability of the proposed $\pifno$ model to real-world traffic data. This further suggests that the $\pifno$ model exhibits similar performance when exclusively trained with a large amount of real-world data, although such an experiment is beyond the scope of this work. Lastly, the above transferability experiment shows the potential of pre-training deep-learning solvers using simulation data and subsequently applying them to scenarios with a limited number of real-world traffic data, as already shown by existing studies \citep{yu2019transfer_tse,thodi2022aniso,thodi2021aniso_itsc}.

\section{Conclusions}
\label{sec:conc}

Physics-informed machine learning methods that seamlessly integrate data and physical models are emerging as popular computational tools for solving traffic and transportation problems. In this paper, we explored an operator learning framework for learning solutions to the first-order macroscopic traffic flow models with an application for predicting and estimating macroscopic traffic density in urban signalized roads. In this framework, a parametric operator is trained to map any input data to its corresponding solution in a supervised learning setting.
The input data can be from heterogeneous, sparse, noisy traffic data sources.
We laid down a generic formulation for the operator learning framework, where two essential elements are the choice of the solution operator and the training data distribution, both of which are problem-dependent. We employ a physics-informed Fourier neural operator (FNO) as the solution operator, where the physics loss function was derived from the discrete conservation law to respect the shock solutions of first-order macroscopic traffic flow models.
We also propose using training data generated using piecewise constant input data to systematically capture the shock and rarefaction solutions observed in the first-order macroscopic traffic flow models. Finally, we empirically quantify the generalization error growth of the trained operator on previously unseen input data.

In the experiments, we study the initial value problem (ring-road traffic system), boundary value problem (urban road with point sensor measurements), and inverse problem (urban road with vehicle trajectory measurements) for the LWR traffic flow model. 
We found that the average prediction error for all three problem settings is $1.30 {\rm ~vehs/km}$, $1.42 {\rm ~vehs/km}$, and $2.30 {\rm ~vehs/km}$ (respectively), with an overall error less than $2.50\%$.
We observe that the addition of the physics loss function has improved the prediction of shock solutions and learning of long-term density dynamics. We also found that the operator can be trained using simple traffic density dynamics, e.g., consisting of $2-3$ initial vehicle queues and $1-2$ traffic signal cycles, and it can predict density solutions for in-homogeneous initial vehicle queues and $\geq 2$ traffic signal cycles with an acceptable error. 
The extrapolation error grew sub-linearly with higher-order initial and boundary conditions for a suitable choice of the model architecture and training data. 
A key finding from this analysis is the importance of the choice of training data and how it depends on the nature of the solution being learned.

The basic theme of the present work is to study the benefits of deep learning methods as a one-stop solution for various problems encountered in traffic simulation and estimation problems. While the conventional methods are well standardized, each method addresses a particular kind of problem. As an example, numerical methods that solve the LWR model work only when (noiseless) initial and boundary conditions are given. If the input data is sparse such as probe vehicle trajectories or limited sensor measurements, one resorts to optimization methods to get an approximate solution. There is no consensus on how to combine noisy measurements with physical models. Further, optimization methods are computationally expensive if one has to re-do the task whenever the problem input data are changed. The proposed physics-informed FNO provided a unified framework to handle all the above problem settings. 
These deep learning methods have great potential in developing fast and accurate simulators and estimators that seamlessly integrate heterogeneous and noisy traffic measurements.
Our efforts continue to extend this framework to network-wide traffic estimations and non-uniform computational grids.


\section*{Acknowledgments} \label{Ack}
This work was supported in part by the NYUAD Center for Interacting Urban Networks (CITIES), funded by Tamkeen under the NYUAD Research Institute Award CG001, and in part by the NYUAD Research Center on Stability, Instability, and Turbulence (SITE), funded by Tamkeen under the NYUAD Research Institute Award CG002. The views expressed in this article are those of the authors and do not reflect the opinions of CITIES, SITE, or their funding agencies.

\bibliography{references}

\appendix

\section{FNO architecture}
\label{sec:appendix_fno}

See Table \ref{tab:model}

\begin{table}[!tbh]
\centering
\caption{FNO architecture and training configuration used in the experiments, unless otherwise specified}
\label{tab:model}
\resizebox{0.65\columnwidth}{!}{%
\begin{tabular}{@{}lll@{}}
\toprule
 & Parameter & Chosen value/setting \\ \midrule
FNO & Number of Fourier layers $(L)$ & $4$ \\
 & \# of Fourier modes in $x$ $(m_k)$ & $24 ~(48 \%)$ \\
 & \# of Fourier modes in $t$ $(n_k)$ & $128 ~(22 \%)$ \\
 & Size of latent dimension $(d_z)$ & $64$ \\
 & Activation function $(\sigma)$ & ReLU \\
 & Up-lifting operator $(P)$ & 1-layer linear NN with depth $128$ \\
 & Down-lifting operator $(Q)$ & 2-layer relu NN with depths $128$ and $1$ \\ \midrule
Training & Training optimizer & Adam gradient descent with lr-decay $0.5 @ 100$ epochs.  \\
 & Optimizer learning rate & $1e^{-3}$ with step-decay \\
 & Training batch size & $128$ samples \\
 & Number of training epochs & $500$ times \\
 & Number of training samples & $6000$ samples \\ \bottomrule
\end{tabular}%
}
\end{table}

\section{Optimization of regularizer weight $\lambda$ in \eqref{eqn:model_pifno}}
\label{sec:appendix_lambda}

\begin{figure}[!tbh]
    \centering
    \includegraphics[width=0.50\columnwidth]{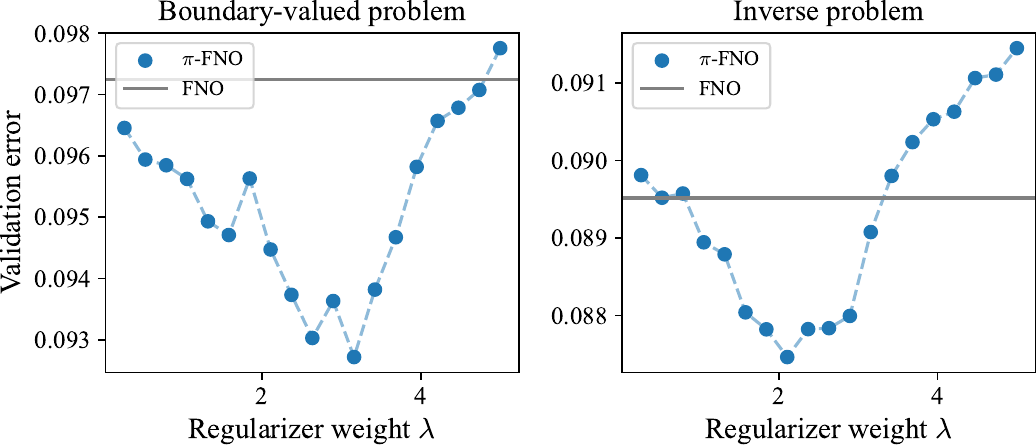}
    \caption{Abc.}
    \label{fig:lambda_opt}
\end{figure}

The regularizer weight $\lambda$ in the physics-informed training \eqref{eqn:model_pifno} is tuned independently. We use a brute-force approach, where we compare the validation error for $\pifno$ solver trained with incremental values for $\lambda$. The results are summarized in Figure \ref{fig:lambda_opt}. The scatter diagram suggests that $\lambda$ values between $2$ and $3$ are a good choice.

\section{Extended results on generalization performance}
\label{sec:appendix_errorrates}

See Table \ref{tab:error_rates_extended}

\begin{table*}[!tbh]
\centering
\caption{Extended results on generalization error rates from Table \ref{tab:error_rates}}
\label{tab:error_rates_extended}
\resizebox{0.98\textwidth}{!}{%
\begin{tabular}{@{}llll@{}}
\toprule
 & Models & Initial condition & Boundary condition \\ \midrule
Model complexity II &M$_5$ $(\mathrm{FNO~depth}~ L=6)$ & $0.13 x_{\alpha}^{0.47} + 1.07$ & $\mathbbm{1}_{x_{\beta} \leq 2.50} \big( 1.15 + 0.05 x_{\beta} \big) + \mathbbm{1}_{x_{\beta} > 2.50} \big( 1.15 + 0.05 x_{\beta} + 0.32 (x_{\beta}-2.50)^{0.80} \big)$ \\
 & M$_6$ $(\mathrm{FNO~depth}~ L=5)$ & $0.09 x_{\alpha}^{0.50} + 1.42$ & $\mathbbm{1}_{x_{\beta} \leq 2.50} \big( 1.41 + 0.15 x_{\beta} \big) + \mathbbm{1}_{x_{\beta} > 2.50} \big( 1.41 + 0.15 x_{\beta} + 0.08 (x_{\beta}-2.50)^{1.52} \big)$ \\
 & M$_7$ $(\mathrm{FNO~depth}~ L=4)$ & $0.12 x_{\alpha}^{0.42} + 1.85$ & $\mathbbm{1}_{x_{\beta} \leq 2.50} \big( 1.96 + 0.08 x_{\beta} \big) + \mathbbm{1}_{x_{\beta} > 2.50} \big( 1.96 + 0.08 x_{\beta} + 0.14 (x_{\beta}-2.50)^{1.36} \big)$ \\
 & M$_8$ $(\mathrm{FNO~depth}~ L=3)$ & $0.12 x_{\alpha}^{0.41} + 1.90$ & $\mathbbm{1}_{x_{\beta} \leq 2.50} \big( 1.93 + 0.14 x_{\beta} \big) + \mathbbm{1}_{x_{\beta} > 2.50} \big( 1.93 + 0.14 x_{\beta} + 0.10 (x_{\beta}-2.50)^{1.42} \big)$ \\
 & M$_9$ $(\mathrm{FNO~depth}~ L=2)$ & $0.15 x_{\alpha}^{0.34} + 3.21$ & $\mathbbm{1}_{x_{\beta} \leq 2.50} \big( 3.11 + 0.24 x_{\beta} \big) + \mathbbm{1}_{x_{\beta} > 2.50} \big( 3.11 + 0.24 x_{\beta} + 0.03 (x_{\beta}-2.50)^{1.74} \big)$ \\\midrule
Training data II & D$_5$ $(\text{Piecewise constant - ours})$ & $0.09 x_{\alpha}^{0.45} + 1.56$ & $\mathbbm{1}_{x_{\beta} \leq 2.50} \big( 1.60 + 0.12 x_{\beta} \big) + \mathbbm{1}_{x_{\beta} > 2.50} \big( 1.60 + 0.12 x_{\beta} + 0.14 (x_{\beta}-2.50)^{1.28} \big)$ \\
 & D$_6$ $(\text{Random})$ & $0.25 x_{\alpha}^{0.75} + 5.35$ & $0.01 x_{\beta}^{2.45} + 3.10$ \\
 & D$_7$ $(\text{Gaussian})$ & $0.25 x_{\alpha}^{0.80} + 3.30$ & $0.05 x_{\beta}^{1.83} + 1.78$ \\
 & D$_8$ $(\text{Sinusoid})$ & $0.25 x_{\alpha}^{0.59} + 2.70$ & $0.08 x_{\beta}^{1.67} + 2.17$ \\ \bottomrule
\end{tabular}%
}
\end{table*}

\end{document}